\documentclass{article}

\PassOptionsToPackage{numbers,sort&compress}{natbib}
\usepackage[preprint]{neurips_2026}

\usepackage[utf8]{inputenc}
\usepackage[T1]{fontenc}
\usepackage{url}
\usepackage{booktabs}
\usepackage{amsfonts}
\usepackage{nicefrac}
\usepackage{microtype}
\usepackage{xcolor}

\usepackage{graphicx}
\usepackage{subcaption}
\usepackage{wrapfig}
\usepackage{amsmath}
\usepackage{amssymb}
\usepackage{mathtools}
\usepackage{amsthm}
\usepackage{hyperref}
\hypersetup{hidelinks}
\usepackage[capitalize,noabbrev]{cleveref}
\usepackage{adjustbox}

\usepackage{bbm}
\usepackage{tabularx}
\usepackage{multirow}
\usepackage{xspace}
\usepackage{colortbl}

\theoremstyle{plain}

\theoremstyle{definition}

\theoremstyle{remark}

\newcommand{\eg}{\emph{e.g.}\xspace}

\definecolor{redbg}{RGB}{235,245,255}
\definecolor{graybg}{RGB}{242,242,242}
\definecolor{smartGreen}{RGB}{119, 185, 85}
\definecolor{smartBlue}{RGB}{85, 140, 220}
\definecolor{smartOrange}{RGB}{235, 145, 80}
\definecolor{histRed}{RGB}{233, 168, 160}
\definecolor{histBlue}{RGB}{166, 206, 227}

\definecolor{lightblue}{HTML}{E6F0FF}
\definecolor{basegray}{gray}{0.92}
\newcolumntype{Y}{>{\centering\arraybackslash}X}
\definecolor{azure}{RGB}{235, 245, 255}
\definecolor{textgray}{RGB}{100, 100, 100}
\definecolor{oracleblue}{RGB}{50, 80, 180}
\definecolor{darkgreen}{rgb}{0.0, 0.5, 0.0}

\title{Dynamic Resolution Routing for Efficient \\ Egocentric Grounding}

\author{%
  \centerline{%
    Huixin Sun$^{1}$,\enspace
    Wangbo Zhao$^{2}$,\enspace
    Fanyue Wei$^{1}$,\enspace
    Qiuxia Lin$^{3}$,\enspace
    Pengzhan Sun$^{1}$,\enspace
    Angela Yao$^{1}$%
  }\\[0.45em]
  {\centerline{$^{1}$National University of Singapore}}\\
  {\centerline{$^{2}$The Hong Kong University of Science and Technology}}\\
  {\centerline{$^{3}$Nanyang Technological University}}
}

\begin{document}

\maketitle

\vspace{-4mm}
\begin{abstract}
Egocentric visual grounding requires high-resolution inputs to localize small objects.
However, scaling Multimodal Large Language Models to this domain is constrained by the excessive cost of visual token processing.
We identify that current efficient strategies based on token reduction are unreliable for selecting object-centric spatial evidence.
To overcome this, we propose \emph{SmartRes}, a framework that performs efficiency optimization in the pixel space via dynamic resolution routing.
SmartRes first encodes a low-resolution view for global context and uses a lightweight router to activate high-resolution patches in object-centric regions and constructs an order‑preserving visual sequence.
To further enable robust routing under severe foreground‑background imbalance, we introduce a margin-regularized routing objective that increases foreground-background logit separation and improves foreground recall.
Experiments on Ego4D and EgoIntention show that SmartRes reduces visual tokens by up to 67\% while retaining 86.4\% of full-resolution performance, and achieves up to $1.66\times$ faster inference than state-of-the-art token reduction methods with higher accuracy.
Furthermore, strong performance on small object grounding indicates the effectiveness of \emph{SmartRes} towards egocentric applications.
Code will be publicly available.
\end{abstract}

\section{Introduction}
\label{sec:intro}
Egocentric grounding~\citep{grauman2022ego4d} aims to localize target objects based on natural-language queries within first-person video streams. This capability is fundamental for understanding human-environment interactions and is useful for applications in robotics, AR/VR, and long-form video understanding.
The latest Multimodal Large Language Models (MLLMs)~\cite{zhang2024llava, bai2023qwen} have impressive visual grounding capabilities, achieving over 90\% accuracy on general benchmarks such as RefCOCO~\cite{yu2016modeling,kazemzadeh2014referitgame}, yet their accuracy remains limited in egocentric scenarios~\citep{sun2025visual}.

\begin{figure}[t]
    \centering
    \includegraphics[width=\linewidth]{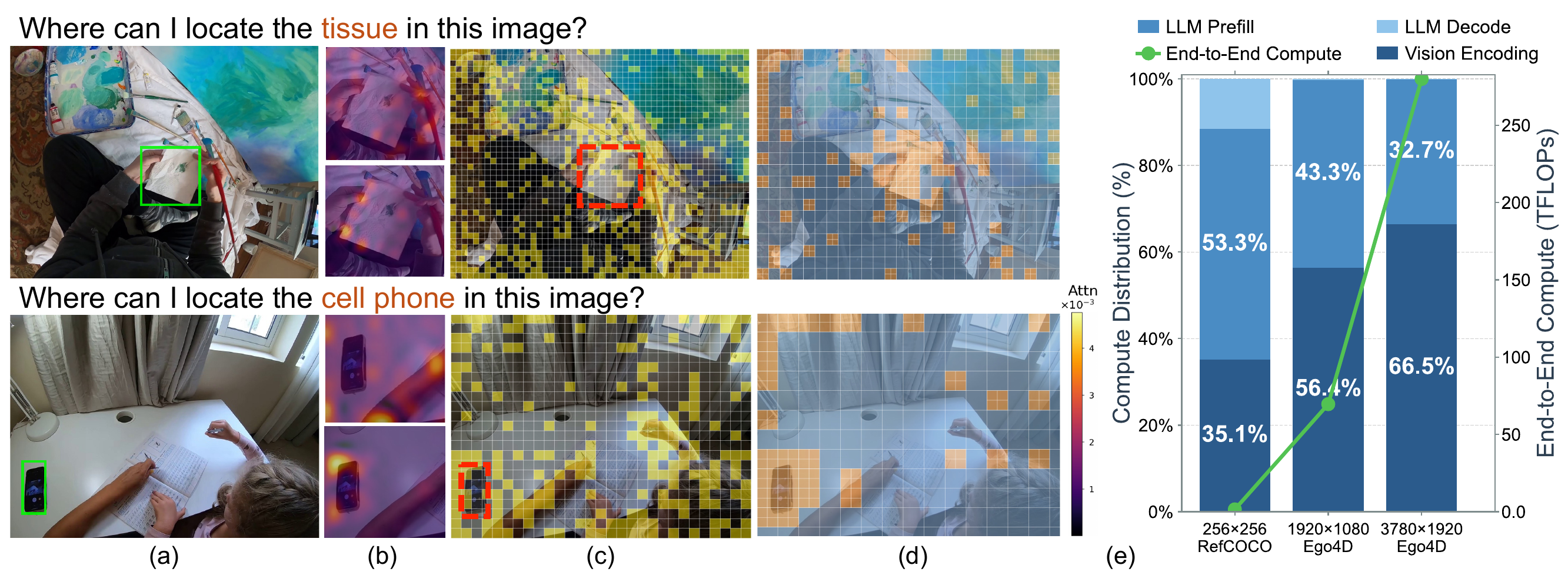}
    \captionsetup{font=small}
    \vspace{-6mm}
    \caption{
    \textbf{Left: Pixel-space dynamic routing \emph{vs.} latent-space token pruning. }
    (a) Input image with the ground-truth target;
    (b) Vanilla cross-attention (top) drifts to artifacts, whereas our method (bottom) calibrates it to the object.
    (c) FastV~\citep{chen2024image} retains tokens 
    (\textcolor[RGB]{255,217,0}{\rule[0.ex]{1.5ex}{1.5ex}}) mostly from the background, failing to preserve fine-grained object evidence. (d) SmartRes allocates high-resolution computation (\textcolor[RGB]{244,177,131}{\rule[0.ex]{1.5ex}{1.5ex}}) to object-centric regions and keeps the remaining visual context compact (\textcolor[RGB]{189,215,238}{\rule[0.ex]{1.5ex}{1.5ex}}).
    \textbf{Right: the visual encoding bottleneck.}
    (e) End-to-end FLOPs (\textcolor[RGB]{0,128,0}{green}) of Qwen2.5-VL-3B-Instruct on RefCOCO and Ego4D frames grow rapidly with input resolution, shifting the compute 
    from LLM prefilling to vision encoding.
    }
    \label{fig:motivation}
    \vspace{-6mm}
\end{figure}

A key challenge in egocentric scenes is that target objects are often small and are observed under rapid viewpoint changes during human interactions~\cite{damen2022rescaling}.
As a result, high-resolution inputs are essential for preserving the fine-grained details required for perception~\cite{sun2019deep}.
However, given that egocentric data are captured at significantly higher and more diverse native resolutions, scaling state-of-the-art MLLMs to this domain becomes prohibitively expensive.
For example, a $3780\times1920$ frame in Any Resolution encoders~\cite{bai2025qwen2} can result in $9.3\mathrm{k}$ visual tokens.
When generating concise text responses that output bounding boxes, the visual encoding cost becomes dominant, consuming up to 66.5\% of the total end-to-end inference budget, as analyzed in Fig.~\ref{fig:motivation}~(e).
This poses a major bottleneck for inference and constrains deployment in edge devices; such a challenge is still underexplored.

Existing efficient methods reduce computation by pruning or merging tokens in the latent space, using heuristics such as decoder attention~\citep{chen2024image,zhang2024sparsevlm}, feature similarity~\citep{bolya2022token}, or activation statistics~\citep{yang2025visionzip}.
However, these heuristics are unreliable for identifying relevant objects in egocentric scenes.
As shown in Fig.~\ref{fig:motivation}~(b), attention can be affected by noisy context, \eg, high-activation artifacts~\citep{darcet2023vision}, or drift to semantically salient but irrelevant regions, such as hands.
This leads to misguided token selection, as illustrated in Fig.~\ref{fig:motivation}~(c), where target object evidence can be discarded while background tokens are retained.
Such methods, as we later show, can severely degrade grounding performance even under moderate token retention ratios.

To avoid using noisy statistical proxies, we shift efficiency optimization from post-hoc token pruning to proactive resolution allocation in the pixel space.
To this end, we introduce \emph{SmartRes}, a framework for dynamic resolution routing at the patch level.
SmartRes utilizes a low-resolution view to provide global spatial guidance, enabling a lightweight router to activate high-resolution patches in object-centric regions, as shown in Fig.~\ref{fig:motivation}~(d).
This yields a variable‑length token sequence that interleaves localized high-resolution evidence with compact low-resolution context.
The router is supervised by box-projected token labels to preserve candidate foreground evidence; however, this is challenged by severe foreground–background imbalance in egocentric data.
To address this, we introduce a margin-regularized routing objective that enforces separation between foreground and background logit distributions, preserving object integrity under a fixed compute budget.
As shown in Fig.~\ref{fig:motivation}~(b), \emph{SmartRes} calibrates visual attention toward object-centric areas via robust foreground selection.
Meanwhile, \emph{SmartRes} reduces the visual encoding costs at its source, and the shortened visual sequence in turn lowers LLM compute.

Experiments on egocentric benchmarks show that \emph{SmartRes} establishes a new Pareto frontier for efficient egocentric grounding.
Specifically, the proposed \emph{SmartRes-Lite} retains 86.4\% of full‑resolution performance using only 33\% of visual tokens, surpassing the competitive down-scaling baseline by a 20.8\% margin.
Furthermore, the performance-oriented \emph{SmartRes-Pro}
retains 89.9\% of full-resolution performance while operating 1.66$\times$ faster than leading pruning methods such as Dyn-LLaVA~\cite{huang2024dynamic}.
The results demonstrate that SmartRes offers a preferable trade-off between fine-grained perception and inference efficiency for egocentric grounding.
Our contributions are summarized as follows:
\begin{itemize}
    \item We observe that latent-space token reduction guided by heuristic
    proxies is unreliable for preserving object-centric spatial evidence in egocentric grounding. To address this, we introduce \emph{SmartRes}, which replaces post-hoc latent pruning with proactive pixel-space selection through dynamic resolution routing.

    \item We introduce a patch-level allocation mechanism where a low-resolution view guides the dynamic activation of high-resolution processing. To ensure robust routing under severe foreground-background imbalance, we introduce a margin-regularized objective that explicitly separates foreground and background logits.

    \item 
    \emph{SmartRes} reduces visual token consumption by up to 67\% while maintaining comparable performance. Notably, it operates 1.66$\times$ faster than leading baselines, establishing a new {Pareto} frontier for efficient grounding on egocentric benchmarks.
\end{itemize}

\section{Related Work}
\label{sec:related}
\noindent\textbf{Visual grounding.}
Visual grounding~\cite{yu2016modeling,mao2016generation,nagaraja2016modeling} aims to localize image regions referred to by natural-language expressions.
Early works mainly studied referring expression comprehension (REC)~\cite{mitchell2013generating,fitzgerald2013learning,kazemzadeh2014referitgame}, focusing on aligning descriptive language with image regions.
Recent MLLMs have improved visual grounding through stronger language-conditioned reasoning~\citep{bai2025qwen3,su2026padt}.
However, localizing small objects remains challenging~\citep{zhang2025mllms} due to inherently limited visual evidence and high sensitivity to localization errors.
This issue is especially pronounced in egocentric visual grounding~\cite{grauman2022ego4d,kurita2023refego,ramanathan2023paco,sun2025visual}, where targets are often small, partially occluded, and surrounded by hands, tools, and cluttered interaction context.
Recent benchmarks (\eg, Ego4D~\cite{grauman2022ego4d}, RefEgo~\cite{kurita2023refego}, and EgoIntention~\cite{sun2025visual}) have driven progress in this setting, primarily emphasizing task formulation, data, and localization accuracy.
Yet efficient inference for high-resolution egocentric grounding remains less explored, especially for small, resolution-sensitive objects.
Our work targets this gap by studying compute-efficient grounding under reduced visual token budgets.

\noindent\textbf{Visual token reduction.}
Recent MLLMs enable high-resolution visual understanding through adaptive resolution strategies~\cite{liu2024llavanext,bai2025qwen2}.
However, the resulting growth in visual tokens introduces substantial memory and latency overhead, motivating visual token reduction methods~\citep{yao2026towards}.
Training-free methods insert lightweight pruning or merging operations into frozen backbones, either in the vision encoder~\citep{bolya2022token,yang2025visionzip,zou2025don,vscan2025} or during LLM prefilling~\citep{chen2024image,xing2024pyramiddrop,zhang2024sparsevlm,bai2025altp}.
Re-training methods, on the other hand, learn explicit compression modules end-to-end~\citep{li2025tokenpacker,huang2025hires,ye2025atpllava,focusui}.
While effective at reducing token sequences, these methods rely on attention, similarity, or learned importance as proxies for spatial utility, which can discard fine-grained spatial details critical for grounding tasks.

\noindent\textbf{Efficient high-resolution visual perception.}
Another line of work focuses on selective high-resolution processing for efficient perception, including crop- or visual-search-based methods~\citep{wu2024vstar,qian2025zoomer}, region-selective sampling~\citep{shi2025ps3,jiang2025teva}, sample-level resolution request~\citep{yang2025visionthink}, and tool-augmented resolution acquisition~\citep{lin2025adaptvision}.
These methods selectively acquire high-resolution evidence through crops, proposed regions, tool invocation, or whole-image resolution requests.
For instance, Zoomer~\citep{qian2025zoomer} builds an efficient image canvas from externally selected crops; TEVA~\citep{jiang2025teva} sparsely samples patches inside externally proposed regions; and AdaptVision~\citep{lin2025adaptvision} learns to invoke a crop tool after low-resolution reasoning.
In contrast, SmartRes learns task-supervised resolution routing at the patch level, enabling object-centric high-resolution coverage without external agents or multi-stage pipelines.

\begin{figure}[t]
    \centering
    \includegraphics[width=\textwidth]{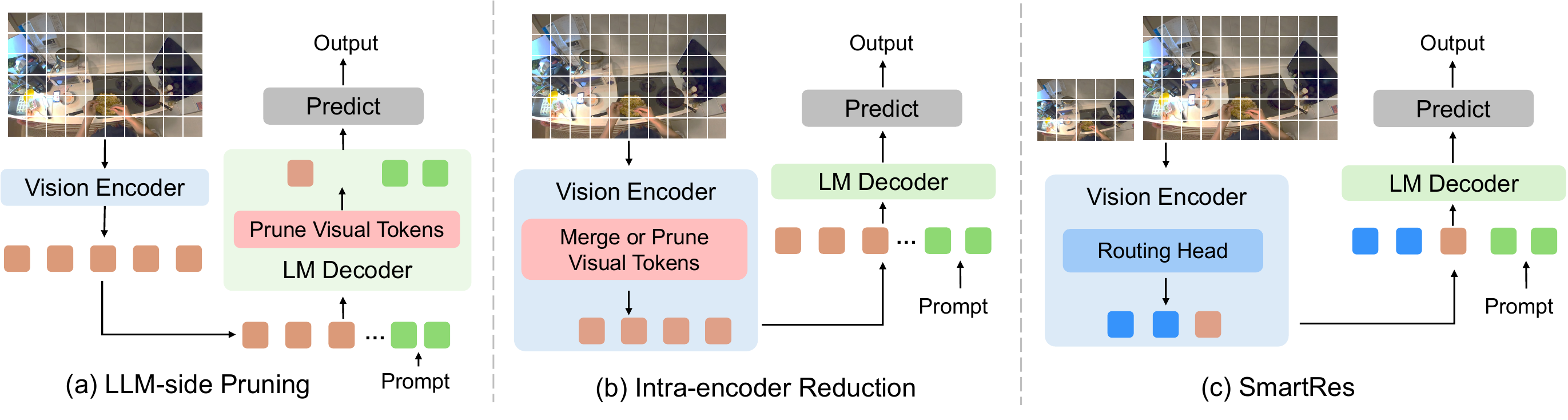}
    \captionsetup{font=small}
    \caption{\textbf{Visual token reduction paradigms.}
    Token types:
    \textcolor{smartGreen}{\rule{1.2ex}{1.2ex}} text,
    \textcolor{smartBlue}{\rule{1.2ex}{1.2ex}} LR visual,
    \textcolor{smartOrange}{\rule{1.2ex}{1.2ex}} HR visual
    (a) {LLM-side pruning} discards tokens with decoder‑side heuristics such as attention scores~\cite{chen2024image}. It lowers LLM compute but has full-resolution visual encoding costs.
    (b) {Intra‑encoder reduction} merges or prunes tokens inside the vision encoder with heuristics such as feature cosine similarity~\cite{bolya2022token}.
    (c) {\emph{SmartRes}} performs low‑resolution guided routing before full high‑resolution encoding, and reduces compute in both the vision encoder and the LLM.
    }
    \label{fig:compared_methods}
    \vspace{-2mm}
\end{figure}

\section{SmartRes: Dynamic Resolution Routing}
\subsection{Preliminaries} \label{sec:preliminaries}

\noindent\textbf{Multimodal Large Language Models (MLLMs).} Modern MLLMs~\cite{liu2023llava,wang2024qwen2} typically adopt an encoder-bridge-decoder architecture. Given an input image $\mathbf{I}\in\mathbb{R}^{H\times W\times 3}$, a vision encoder maps the image into $N_v$ visual features $\mathbf{V}\in\mathbb{R}^{N_v\times D_v}$. Concurrently, the input text is tokenized and transformed into text embeddings $\mathbf{T}\in\mathbb{R}^{N_t\times D}$ via the language model's embedding layer. To bridge the modality gap, visual features are aligned with the language model's latent dimension $D$ using a linear layer $\mathrm{Proj}:\mathbb{R}^{D_v}\!\rightarrow\!\mathbb{R}^{D}$. The resulting visual and textual tokens are concatenated and fed into the language model (LM):
\begin{equation}
\label{eq:mllm_inference}
\mathbf{y}_t =
\begin{cases}
\mathrm{LM}_{\mathrm{prefilling}}
\left([\mathrm{Proj}(\mathbf{V}); \mathbf{T}]\right),
& t=1, \\[2pt]
\mathrm{LM}_{\mathrm{decoding}}
\left([\mathbf{C}_{t-1}; \mathbf{y}_{1:t-1}]\right),
& t>1.
\end{cases}
\end{equation}
where $\mathbf{y}_t$ denotes the $t$-th generated token. In the prefilling stage ($t=1$), the model processes the entire multimodal prefix in parallel to compute the initial KV cache. For $t > 1$, the model enters the auto-regressive decoding stage, where $\mathbf{C}_{t-1}$ represents the cached KV pairs from previous steps.

\noindent\textbf{MLLMs for Egocentric Grounding and Its Compute Bottleneck.}
Given a first-person image $\mathbf{I}$ and a grounding instruction $\mathbf{T}$ , the inference process is formulated as:
\begin{equation}
    \mathbf{B} = \operatorname{MLLM}(\mathbf{I}, \mathbf{T})
\label{eq:ego}
\end{equation}
where $\mathbf{B}$ is a coordinate tuple $[x_{\min}, y_{\min}, x_{\max}, y_{\max}]$ representing the bounding box of the target object.
In this work, we identify a \emph{fundamental tension}  between the high-resolution requirements of egocentric grounding to preserve fine-grained spatial details and the resolution-dependent cost of visual processing.
As shown in Fig.~\ref{fig:motivation}\,(e), high-resolution inference is bottlenecked by vision encoding and LLM prefilling.
Scaling the input from $256\!\times\!256$ to $3780\!\times\!1920$, a resolution commonly encountered in egocentric scenarios, increases the vision-encoding cost from $0.53$T to $186.03$T FLOPs and the LLM-prefilling cost from $0.81$T to $93.70$T FLOPs.
In contrast, the autoregressive decoding remains lightweight as grounding outputs are short  (typically $<$350).
Existing token pruning (Fig.~\ref{fig:compared_methods}\,(a)) and token merging (Fig.~\ref{fig:compared_methods}\,(b)) methods reduce the visual sequence length before or inside the LLM but cannot tackle the dominant visual encoding cost.
\subsection{SmartRes Architecture}
\label{subsec:smartres_architecture}
\begin{figure*}[t]
    \centering
    \includegraphics[width=0.98\textwidth]{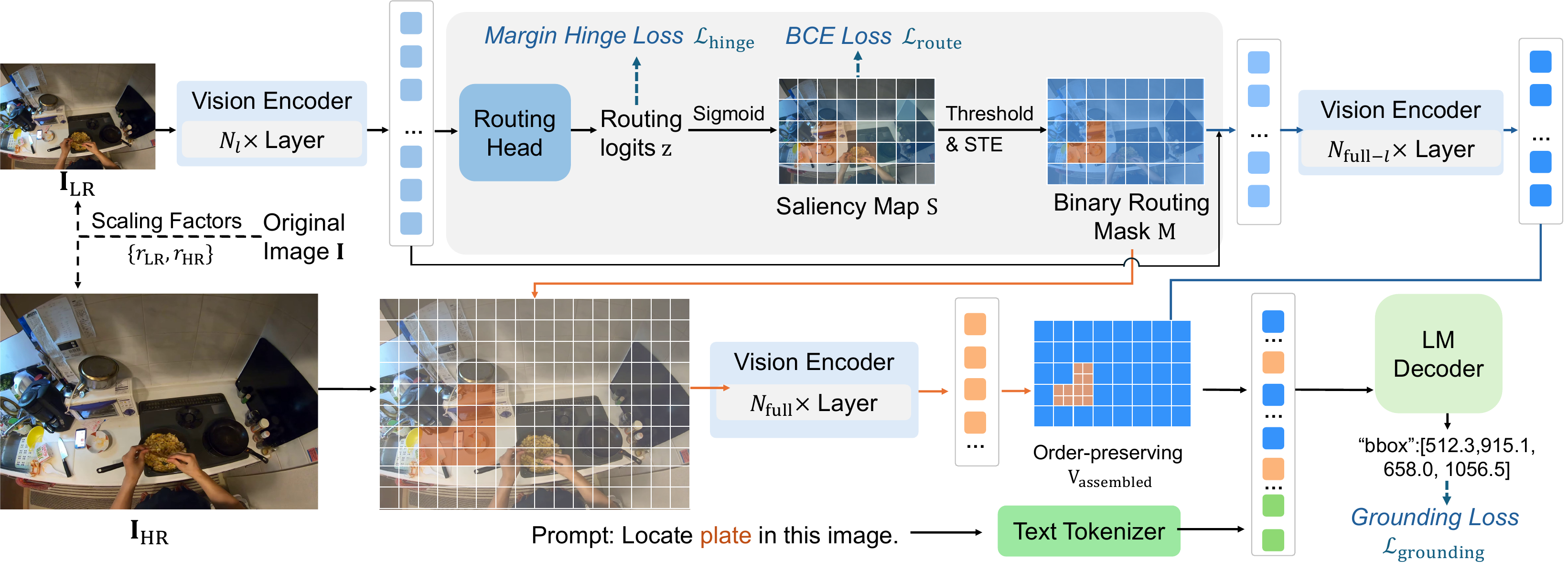}
    \captionsetup{font=small}
    \caption{
    \textbf{Overview of the SmartRes framework.}
    Token types:
    \textcolor[RGB]{176,215,246}{\rule[0.ex]{1.5ex}{1.5ex}}~LR feature $\mathbf{F}$,
    \textcolor[RGB]{68,114,196}{\rule[0.ex]{1.5ex}{1.5ex}}~LR visual feature $\mathbf{V}_{\text{LR}}$,
    \textcolor[RGB]{244,177,131}{\rule[0.ex]{1.5ex}{1.5ex}}~HR visual feature $\mathbf{V}_{\text{HR}}$,
    \textcolor[RGB]{169,209,142}{\rule[0.ex]{1.5ex}{1.5ex}}~textual token.
    SmartRes first processes a low-resolution input $\mathbf{I}_{\text{LR}}$ to extract routing features $\mathbf{F}$.
    A lightweight routing head predicts a saliency map from these features, supervised by the routing loss $\mathcal{L}_{\text{route}}$ and the margin hinge loss $\mathcal{L}_{\text{hinge}}$.
    The Straight-Through Estimator (STE) binarizes the saliency map into a routing mask $\mathbf{M}$, which activates selected grounding-relevant patches from the high-resolution input $\mathbf{I}_{\text{HR}}$.
    The selected HR features are assembled with the compact LR context in raster-scan order, yielding a routed visual sequence passed to the LM decoder.
    }
    \label{fig:architecture}
    \vspace{-3mm}
\end{figure*}

To resolve the above tension, we propose \emph{SmartRes}, a dynamic resolution routing framework.
As shown in Fig.~\ref{fig:architecture}, SmartRes encodes a low-resolution view to obtain global spatial guidance, learns a patch-level router to activate high-resolution evidence, and assembles the selected HR features with compact LR context in raster-scan order.
This design reduces visual processing cost at its source while preserving localized details needed for accurate grounding.
We next describe the image preparation, learnable routing strategy, and visual token assembly.

\noindent\textbf{Image preparation.}
From the original image $\mathbf{I}$, we derive two aligned views: a low-resolution image $\mathbf{I}_{\mathrm{LR}}$ for routing and a high-resolution image $\mathbf{I}_{\mathrm{HR}}$ for selective encoding.
For target token ratios $r_{\mathrm{LR}},r_{\mathrm{HR}}\in(0,1]$, the LR and HR views are resized by isotropically scaling both height and width by $\sqrt{r_{\mathrm{LR}}}$ and $\sqrt{r_{\mathrm{HR}}}$, respectively, preserving the original aspect ratio.
We then snap the resized side lengths to multiples of the effective visual-token stride $F=PM$, where $P$ is the patch size and $M$ is the spatial merge size.
This defines aligned LR and HR routing grids and induces a HR-to-LR token correspondence; see Sec.~\ref{supp:processing} of the \textit{Supplementary}.
The LR view $\mathbf{I}_{\mathrm{LR}}$ is then passed through the vision encoder, and its layer-$\ell$ features
$\mathbf F\in\mathbb R^{N_{\mathrm{LR}}\times D_{\mathrm v}}$ are used as input to the routing module.

\noindent\textbf{Learnable patch-level resolution routing.}
High-resolution evidence is essential for grounding small objects, yet uniform high-resolution encoding is computationally prohibitive.
SmartRes addresses this trade-off with a learnable patch-level routing policy that predicts where additional HR evidence should be allocated.
Specifically, we introduce a lightweight MLP-based routing head to predict a routing saliency map
$\mathbf{S}$ from the LR features $\mathbf{F}$:
\begin{equation}
\label{eq:saliency_score}
\mathbf{z}=\mathrm{MLP}(\mathbf{F})\in\mathbb{R}^{N_{\mathrm{LR}}},\qquad
\mathbf{S}=\sigma(\mathbf{z})\in[0,1]^{N_{\mathrm{LR}}},
\end{equation}
where $N_{\mathrm{LR}}$ denotes the number of LR locations, $\mathbf{z}=\{z_i\}_{i=1}^{N_{\mathrm{LR}}}$ are routing logits, and $\sigma(\cdot)$ is the element-wise sigmoid function.
This saliency map is converted into a binary routing mask $\mathbf{M}$ by thresholding $\mathbf{S}$ at a hyperparameter $\tau$:
\begin{equation}
\label{eq:Mvis}
\mathbf{M}
=
\mathrm{STE}(\mathbf{S}>\tau)
\in \{0,1\}^{N_{\mathrm{LR}}},
\end{equation}
where $\mathrm{STE}(\cdot)$ denotes the Straight-Through Estimator~\cite{bengio2013estimating}.
During the forward pass, a hard binary threshold is applied; in the backward pass, gradients are directly propagated from $\mathbf{M}$ to $\mathbf{S}$, enabling end-to-end optimization of routing decisions.
A selected location, $\mathbf M_i=1$, indicates that the LR evidence at location $i$ is expected to benefit from additional HR detail and activates its corresponding HR patch group; otherwise, the compact LR feature is retained.
In this way, SmartRes converts dense high-resolution encoding into a sparse, learnable patch activation procedure, recovering high-resolution local details where needed while keeping the remaining visual context compact.

\begin{wrapfigure}{r}{0.35\linewidth}
    \centering
    \includegraphics[width=\linewidth]{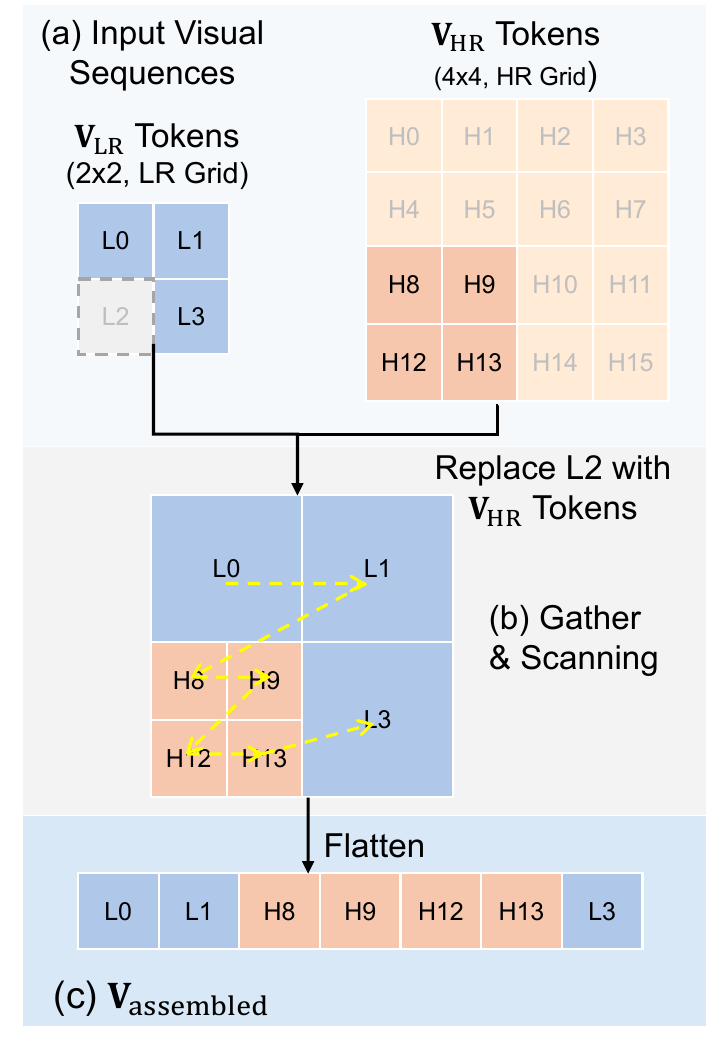}
    \vspace{-5mm}
    \captionsetup{font=scriptsize}
    \caption{
    Selected LR locations, \eg, L2, are replaced in place by their corresponding contiguous HR token groups, \eg, H8, H9, H12, and H13, while unselected locations remain as LR context. }
    \label{fig:multi-res-token}
    \vspace{-2mm}
\end{wrapfigure}
\noindent\textbf{Order-preserving visual token assembly.}
The routing mask $\mathbf M$ determines LR retention and HR replacement.
After routing at layer $\ell$, the LR features $\mathbf F$ are propagated through the remaining vision encoder blocks to obtain the low-resolution visual features
$\mathbf V_{\mathrm{LR}}\in\mathbb R^{N_{\mathrm{LR}}\times D_{\mathrm v}}$.
The routing mask $\mathbf M$ is then upsampled to align with the high-resolution image.
Based on the HR$\!\rightarrow$LR correspondence from Eq.~\eqref{eq:hr2lr_snap}, selected patches from $\mathbf I_{\mathrm{HR}}$ are processed through the vision encoder to obtain HR visual features
$\mathbf{V}_{\text{HR}} \in \mathbb{R}^{N'_{\text{HR}} \times D_v}$, where $N'_{\text{HR}}$ denotes the number of selected HR patches.
The assembled visual sequence interleaves LR context with HR details:
\begin{equation}
\label{eq:Ems}
\mathbf{V}_{\mathrm{assembled}} = \mathcal{G}\left(\mathbf{V}_{\mathrm{LR}}, \mathbf{V}_{\mathrm{HR}}, \mathbf{M}\right).
\end{equation}
A key property of $\mathcal{G}(\cdot)$ is that it preserves the canonical LR raster-scan order. Unselected LR locations remain as compact patch tokens, while selected LR locations are expanded in place into their corresponding contiguous HR token groups; see Fig.~\ref{fig:multi-res-token}.
$\mathbf V_{\mathrm{assembled}}$ is zero-padded to ensure divisibility by the spatial merge unit $M^2$, and is then passed through $\mathrm{Proj}$ to yield the final visual token sequence $\mathbf{V}_{\mathrm{final}}\in\mathbb{R}^{N_{\mathrm{final}}\times D}$.
Thus, $\mathbf V_{\mathrm{final}}$ covers the entire image with a variable number of tokens without disrupting the spatial order.
\subsection{Routing Optimization with Margin Regularization}
\label{sec:cfcl}
The routing strategy is trained end-to-end under the primary grounding task while receiving object-centric supervision.
Specifically, we encourage high saliency scores on tokens overlapping the target box and suppress task-irrelevant background tokens.
Given the LR image $\mathbf{I}_{\mathrm{LR}}$ and the ground-truth bounding box $\hat{\mathbf{B}}$, we rasterize $\hat{\mathbf{B}}$ onto the LR patch grid to obtain token-wise labels
$\hat{\mathbf{M}}\in\{0,1\}^{N_{\mathrm{LR}}}$, and supervise the predicted saliency scores $\mathbf{S}$ using a standard binary cross-entropy (BCE) loss:
\begin{equation}
\label{eq:Lconf}
\mathcal{L}_{\mathrm{route}}
=
\frac{1}{N_{\mathrm{LR}}}
\sum_{i=1}^{N_{\mathrm{LR}}}
\operatorname{BCE}
\left(
\mathbf{S}[i],\hat{\mathbf{M}}[i]
\right).
\end{equation}

\begin{wrapfigure}{r}{0.48\linewidth}
    \centering
    \vspace{-1.5mm}
    \includegraphics[width=0.98\linewidth]{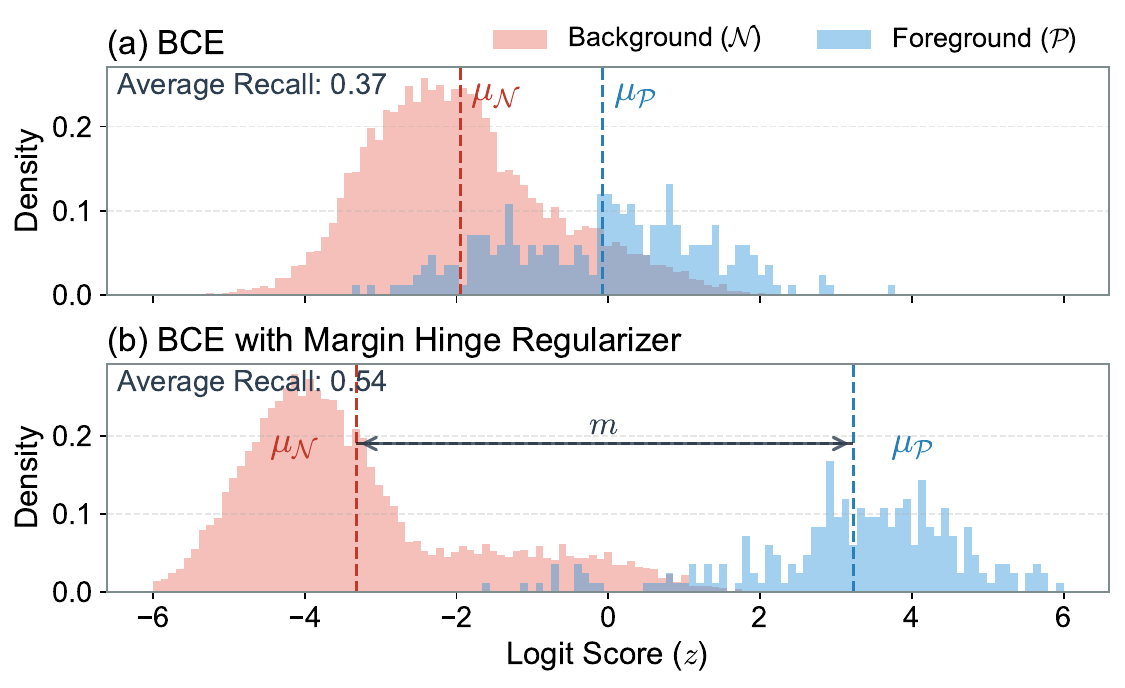}
    \captionsetup{font=small}
    \caption{\textbf{Routing logit separation.}
    Routing logits of 3{,}000 randomly sampled tokens from the Ego4D test set.
    (a) Token-wise BCE yields highly overlapping foreground/background distributions and low foreground recall ($R=0.37$).
    (b) Adding $\mathcal{L}_{\mathrm{hinge}}$ increases logit separation and improves foreground recall ($R=0.54$).}
    \label{fig:contrastive_mechanism}
\end{wrapfigure}

\noindent\textbf{Margin hinge regularization.}
However, we observe that supervising the routing head solely with the BCE loss often yields suboptimal saliency maps.
This result is primarily attributed to the extreme foreground-background imbalance inherent in egocentric grounding data: target objects typically occupy a minute fraction of the spatial area, while the vast majority of visual tokens represent background context. Consequently, the per-token routing loss is dominated by an abundance of easily classified background tokens. This imbalance dilutes the gradients essential for learning sparse foreground features,  resulting in reduced routing recall and overlapping logit distributions between foreground and background regions, as illustrated in Fig.~\ref{fig:contrastive_mechanism} (a).
To address this imbalance, we introduce a margin-based hinge regularizer on the routing logits to encourage class-level separation.
Let $\mathcal{P}=\{i\mid \hat{M}_i=1\}$ and $\mathcal{N}=\{i\mid \hat{M}_i=0\}$ denote foreground and background token sets.
Given routing logits $\mathbf z=\{z_i\}_{i=1}^{N_{\mathrm{LR}}}$, we define the class-wise logit means as
$\mu_{\mathcal P}=\frac{1}{|\mathcal P|}\sum_{i\in\mathcal P}z_i$
and
$\mu_{\mathcal N}=\frac{1}{|\mathcal N|}\sum_{i\in\mathcal N}z_i$,
and encourage a margin $m$ between them:
\begin{equation}
\label{eq:Lhinge}
\mathcal{L}_{\mathrm{hinge}}
=
\left[
m-\left(\mu_{\mathcal{P}}-\mu_{\mathcal{N}}\right)
\right]_+,
\end{equation}
where $[\,\cdot\,]_+=\max(0,\cdot)$.
As shown in Fig.~\ref{fig:contrastive_mechanism}(b), the hinge term enlarges the foreground--background logit gap and improves foreground recall under spatial sparsity.
Gradient analysis and comparisons with alternative imbalance-aware objectives, including Dice loss~\citep{milletari2016vnet}, focal loss~\citep{lin2017focal}, and Tversky loss~\citep{salehi2017tversky}, are provided in Sec.~\ref{app:imbalance_losses} of the \textit{Supplementary}.

\noindent\textbf{Optimization objective.}
\label{sec:optimization_objective}
The overall training objective combines the primary grounding loss with the routing supervision terms.
For the grounding task, we optimize the standard next-token cross-entropy loss over the ground-truth output sequence
$\hat{\mathbf y}=(\hat y_1,\ldots,\hat y_L)$,
which is tokenized from the target bounding box $\hat{\mathbf B}$, using teacher forcing:
\begin{equation}
\label{eq:Ltask_ce}
\mathcal{L}_{\mathrm{grounding}}
=
-\sum_{t=1}^{L}
\log P\!\left(
\hat y_t
\mid
\mathbf V_{\mathrm{final}}, \mathbf T, \hat{\mathbf y}_{1:t-1}
\right),
\end{equation}
where $\mathbf T$ denotes the grounding instruction, $\mathbf V_{\mathrm{final}}$ denotes the routed visual token sequence, and $P(\cdot)$ is the probability distribution predicted by the model.
The final objective is:
\begin{equation}
\label{eq:Ltotal}
\mathcal{L}_{\mathrm{total}}
=
\mathcal{L}_{\mathrm{grounding}}
+
\lambda_{\mathrm{route}}\mathcal{L}_{\mathrm{route}}
+
\lambda_{\mathrm{hinge}}\mathcal{L}_{\mathrm{hinge}},
\end{equation}
where $\lambda_{\mathrm{route}}$ and $\lambda_{\mathrm{hinge}}$ are weighting hyperparameters.

\begin{table*}[t]
  \centering
  \setlength{\tabcolsep}{2.4pt}
  \renewcommand{\arraystretch}{1.06}
  \caption{\textbf{Main results on egocentric grounding benchmarks.}
  Vanilla denotes Qwen2.5-VL-3B-Instruct.
  EgoInt-C/U denote the Context/Uncommon subsets of EgoIntention.
  Overall is the mean P$@0.5$ over Ego4D, EgoInt-C, and EgoInt-U.
  Avg. denotes the mean performance retention rate across all 9 metrics relative to Vanilla.
  \textbf{Bold} marks the best value among efficient methods.}
  \label{tab:main_results}
  \begin{adjustbox}{max width=\textwidth}
  \begin{tabular}{@{} l c rrr rrr rrr r rrr @{}}
  \toprule
  \multirow{2}{*}{\textbf{Method}}
  & \multirow{2}{*}{\textbf{Ratio} $\downarrow$}
  & \multicolumn{3}{c}{\textbf{Ego4D}}
  & \multicolumn{3}{c}{\textbf{EgoInt-C}}
  & \multicolumn{3}{c}{\textbf{EgoInt-U}}
  & \multicolumn{1}{c}{\textbf{Overall}}
  & \multicolumn{1}{c}{\textbf{Avg.} $\uparrow$}
  & \multicolumn{1}{c}{\textbf{FLOPs} $\downarrow$}
  & \multicolumn{1}{c}{\textbf{Lat.} $\downarrow$} \\
  \cmidrule(lr){3-5}
  \cmidrule(lr){6-8}
  \cmidrule(lr){9-11}
  \cmidrule(lr){12-12}
  \cmidrule(lr){13-13}
  \cmidrule(lr){14-14}
  \cmidrule(lr){15-15}
  & &
  P$@0.5$ & P$@0.3$ & mIoU
  & P$@0.5$ & P$@0.3$ & mIoU
  & P$@0.5$ & P$@0.3$ & mIoU
  & P$@0.5$ & (\%) & (T) & (ms) \\
  \midrule
  \rowcolor{basegray}
  Vanilla & 100\%
  & 63.22 & 68.51 & 57.56
  & 60.01 & 65.15 & 54.62
  & 54.77 & 59.18 & 49.98
  & 59.33 & 100.0 & 11.32 & 553.6 \\
  \midrule
  \multirow{3}{*}{Down-scaling}
  & 10\%
  & 21.34 & 33.26 & 21.85
  & 20.65 & 27.05 & 20.75
  & 19.50 & 23.73 & 19.60
  & 20.50 & 38.8 & \textbf{1.55} & \textbf{139.6} \\
  & 32\%
  & 32.76 & 43.12 & 30.92
  & 31.83 & 37.61 & 31.01
  & 23.10 & 27.69 & 22.69
  & 29.23 & 52.3 & 3.94 & 240.8 \\
  & 50\%
  & 52.17 & 56.08 & 47.19
  & 50.25 & 53.77 & 46.90
  & 42.70 & 46.02 & 40.91
  & 48.37 & 81.8 & 5.89 & 323.6 \\
  \midrule
  ToMe~\citep{bolya2022token} & 70\%
  & 22.83 & 35.66 & 25.12
  & 20.36 & 33.99 & 22.85
  & 17.96 & 29.65 & 20.09
  & 20.38 & 42.5 & 8.52 & 408.8 \\
  VisionZip~\citep{yang2025visionzip} & 70\%
  & 34.18 & 42.63 & 31.04
  & 30.73 & 38.05 & 27.62
  & 28.02 & 34.56 & 25.33
  & 30.98 & 54.5 & 8.63 & 305.7 \\
  FastV~\citep{chen2024image} & 70\%
  & 51.65 & 60.29 & 47.03
  & 50.33 & 57.58 & 45.30
  & 43.61 & 49.83 & 39.44
  & 48.53 & 83.3 & 8.92 & 495.5 \\
  Dyn-LLaVA~\citep{huang2024dynamic} & 70\%
  & 54.42 & 60.30 & 48.24
  & 52.12 & 58.20 & 48.15
  & 44.54 & 47.65 & 40.86
  & 50.36 & 85.1 & 9.16 & 506.7 \\
  \midrule
  ToMe~\citep{bolya2022token} & 50\%
  & 11.41 & 22.62 & 15.05
  & 10.78 & 21.36 & 14.41
  & 9.74 & 19.40 & 13.08
  & 10.64 & 25.7 & 7.42 & 365.5 \\
  VisionZip~\citep{yang2025visionzip} & 50\%
  & 20.28 & 31.64 & 19.74
  & 18.87 & 26.23 & 18.53
  & 17.22 & 23.82 & 16.96
  & 18.79 & 36.0 & 7.98 & 293.4 \\
  FastV~\citep{chen2024image} & 50\%
  & 45.52 & 53.00 & 41.44
  & 47.32 & 54.68 & 42.65
  & 35.74 & 40.90 & 32.85
  & 42.86 & 73.6 & 7.15 & 410.2 \\
  Dyn-LLaVA~\citep{huang2024dynamic} & 50\%
  & 52.48 & 61.15 & 48.72
  & 50.76 & 57.38 & 46.65
  & 39.19 & 42.87 & 36.74
  & 47.48 & 81.4 & 7.41 & 419.3 \\
  \midrule
  \multicolumn{15}{@{}l}{\textit{Adaptive: 10\% $\rightarrow$ 50\% tokens}} \\
  \rowcolor{lightblue}
  \textbf{SmartRes-Lite} & 33\%
  & 54.55 & 61.46 & 48.42
  & 52.15 & 58.25 & 46.39
  & 46.63 & 51.94 & 41.71
  & 51.11 & 86.4 & 4.05 & 252.6 \\
  \midrule
  \multicolumn{15}{@{}l}{\textit{Adaptive: 10\% $\rightarrow$ 100\% tokens}} \\
  \rowcolor{lightblue}
  \textbf{SmartRes-Pro} & 55\%
  & \textbf{55.45} & \textbf{62.18} & \textbf{49.10}
  & \textbf{53.84} & \textbf{60.81} & \textbf{47.48}
  & \textbf{50.04} & \textbf{56.37} & \textbf{44.36}
  & \textbf{53.11} & \textbf{89.9} & 6.58 & 365.8 \\
  \bottomrule
  \end{tabular}
  \end{adjustbox}
  \vspace{-2mm}
  \end{table*}
  \section{Experiments}
  \label{sec:experiments}
  \subsection{Benchmarks and Implementation Details}
  \noindent\textbf{Datasets}.
  We evaluate
  \emph{SmartRes} on public egocentric visual grounding benchmarks, {Ego4D}~\cite{grauman2022ego4d} and {EgoIntention}~\cite{sun2025visual}.
  These egocentric datasets contain diverse high-resolution frames and a
  larger
  proportion of small objects which require fine-grained visual details to localize; see Tab.~\ref{tab:dataset_stats}.
  We additionally evaluate \emph{SmartRes} on the standard referring expression comprehension (REC) benchmarks {RefCOCO}, {RefCOCO+}, and {RefCOCOg}~\cite{kazemzadeh2014referitgame,yu2016modeling,mao2016generation}.
  \\
\noindent\textbf{Metrics.}
Grounding performance is evaluated by Precision at IoU 0.3/0.5 (P$@0.3$, P$@0.5$) and mean IoU (mIoU).
Following~\cite{zhang2025mllms}, objects are grouped by relative box area $S$ into small ($S<0.005$), medium ($0.005\leq S<0.05$), and large ($S\geq0.05$), with $\mathrm{P}_{\mathrm{s}}$, $\mathrm{P}_{\mathrm{m}}$, and $\mathrm{P}_{\mathrm{l}}$ denoting scale-specific P$@0.5$.
For efficiency, we compute LLM prefilling FLOPs and GPU latency with synchronized \texttt{torch.cuda.Event} timing following~\cite{zhang2024sparsevlm}. \\
\noindent\textbf{Compared Methods.}
We evaluate in two adaptive configurations:
\emph{SmartRes-Lite} ($r_{\text{LR}}$=10\%, $r_{\text{HR}}$=50\% budget) and \emph{SmartRes-Pro} ($r_{\text{LR}}$=10\%, $r_{\text{HR}}$=100\% budget).
We compare against representative state-of-the-art efficiency strategies:
(1) \textit{Token pruning} (VisionZip~\cite{yang2025visionzip}, FastV~\cite{chen2024image}, Dyn-LLaVA~\citep{huang2024dynamic}, VScan~\citep{vscan2025});
(2) \textit{Token merging} (ToMe~\cite{bolya2022token});
and (3) \textit{Naive down-sampling} baselines, where
images are uniformly resized
to a target token budget. \\
\noindent\textbf{Implementation Details.}
For fair comparison, we reproduce all baselines and comparisons with \texttt{Qwen2.5-VL} base model and fine-tune
with the same LoRA rank.
We fine-tune for 3 epochs with learning rate $1\times10^{-4}$ using a cosine schedule.
For \emph{SmartRes}, the router utilizes features from layer $l=30$ and is trained with a learning rate of $5\times10^{-4}$ and threshold $\tau = 0.5$. All experiments are conducted with 2 NVIDIA H100‑96GB GPUs for training and one H100-96GB GPU for inference.
\vspace{-2mm}
\subsection{Main Experiments on Egocentric Benchmarks}
\label{sec:main_results}
\vspace{-1mm}
Tab.~\ref{tab:main_results} summarizes grounding performance and computational efficiency on Ego4D and EgoIntention across methods.
As illustrated, \emph{SmartRes} achieves the Pareto-optimal accuracy-efficiency trade-offs.
\vspace{-2mm}
\begin{figure}[ht]
  \centering
  \includegraphics[width=\textwidth]{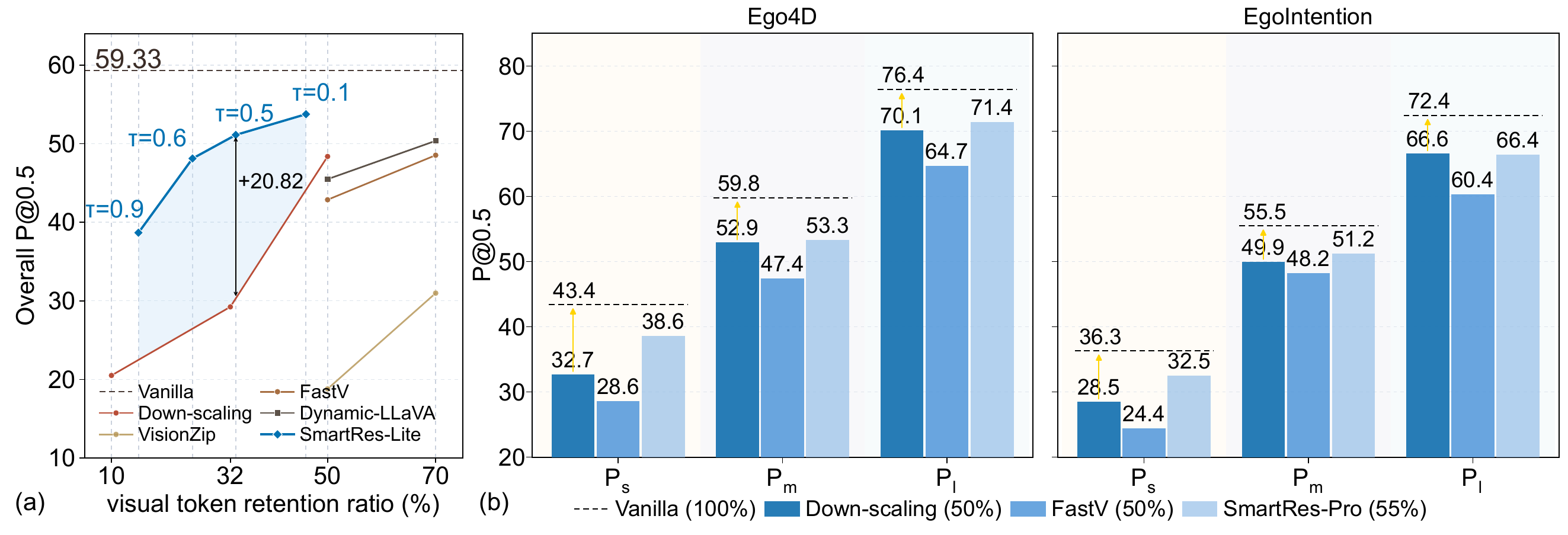}
  \caption{
  \textbf{(a) Adaptive compute allocation during inference.} Varying the routing threshold $\tau$ in \emph{SmartRes-Lite} yields different token retention ratios, establishing an adaptive accuracy--efficiency frontier that consistently outperforms uniform down-scaling at comparable token budgets.
  \textbf{(b) Performance by object scale.} \emph{SmartRes-Pro} better preserves small object performance by routing high-resolution computation to object-centric regions.
 }
  \label{fig:results}
  \vspace{-2mm}
\end{figure}

\noindent\textbf{Grounding Performance.}
We achieve high grounding accuracy under strong compression.
\emph{SmartRes-Lite} retains 86.4\% of full-resolution performance using only 33\% of visual tokens on average, surpassing uniform down-scaling at a comparable budget by 34.1\% in retention and improving Overall P$@0.5$ from 29.23 to 51.11.
Fig.~\ref{fig:results}~(a) further shows that \emph{SmartRes-Lite} consistently outperforms uniform down-scaling across different routing thresholds $\tau$, forming an adaptive accuracy--efficiency frontier.
\emph{SmartRes-Pro} improves retention to 89.9\% at a 55\% token budget and achieves the best grounding accuracy among efficient methods.
In contrast, token-reduction methods degrade notably under compression.
At 70\% tokens, ToMe and VisionZip retain only 42.5\% and 54.5\%, while FastV and Dyn-LLaVA retain 83.3\% and 85.1\% but require higher FLOPs and latency.
At 50\% tokens, the strongest comparison, Dyn-LLaVA, drops to 81.4\% retention.
These results support our observation that latent-space pruning can discard fine-grained spatial evidence needed for accurate localization.
\noindent\textbf{Efficiency.}
The uniform down-sampling baselines have the lowest compute but also suffer from severe performance degradation.
SmartRes achieves a better trade-off, incurring negligible overhead over down-sampling while significantly outperforming token pruning.
\emph{SmartRes-Lite} reduces FLOPs by 64\% and latency by 54\% compared to the full-resolution Vanilla model.
Crucially, it operates 1.66$\times$ faster than the 50\%-token Dyn-LLaVA variant while delivering 5.0\% higher retention.
\begin{table*}[t]
\centering
\footnotesize
\setlength{\tabcolsep}{2.4pt}
\renewcommand{\arraystretch}{1.02}
\caption{\textbf{Main results on referring expression comprehension (REC).}
Vanilla denotes Qwen2.5-VL-3B-Instruct.
We report P$@0.5$ on RefCOCO, RefCOCO+, and RefCOCOg.
Overall is the arithmetic mean over the eight evaluation splits.
Avg. denotes mean performance retention over the eight splits relative to Vanilla.
\textbf{Bold} marks the best value among efficient methods.}
\vspace{-2mm}
\label{tab:refcoco_main}
\begin{tabularx}{\textwidth}{@{} l c *{3}{Y} *{3}{Y} *{2}{Y} Y Y @{}}
\toprule
\multirow{2}{*}{\textbf{Method}}
& \multirow{2}{*}{\textbf{Ratio} $\downarrow$}
& \multicolumn{3}{c}{\textbf{RefCOCO}}
& \multicolumn{3}{c}{\textbf{RefCOCO+}}
& \multicolumn{2}{c}{\textbf{RefCOCOg}}
& \multicolumn{1}{c}{\textbf{Overall}}
& \multicolumn{1}{c}{\textbf{Avg.} $\uparrow$} \\
\cmidrule(lr){3-5}
\cmidrule(lr){6-8}
\cmidrule(lr){9-10}
\cmidrule(lr){11-11}
\cmidrule(lr){12-12}
& &
Val & TestA & TestB
& Val & TestA & TestB
& Val & Test
& P$@0.5$ & (\%) \\
\midrule
\rowcolor{basegray}
Vanilla & 100\%
& 91.19 & 92.88 & 86.85
& 86.15 & 90.02 & 80.41
& 87.96 & 88.12
& 87.95 & 100.0 \\
\midrule
\multirow{3}{*}{Down-scaling}
& 50\%
& 66.92 & 68.74 & 63.85
& 63.22 & 66.62 & 59.12
& 64.55 & 65.22
& 64.78 & 73.7 \\
& 32\%
& 57.86 & 56.06 & 52.42
& 54.66 & 54.33 & 48.53
& 55.81 & 53.19
& 54.11 & 61.5 \\
& 10\%
& 41.47 & 36.61 & 38.56
& 39.18 & 35.48 & 35.70
& 40.00 & 34.73
& 37.72 & 42.9 \\
\midrule
FastV~\citep{chen2024image} & 50\%
& 61.02 & 62.71 & 59.25
& 55.40 & 58.62 & 51.72
& 58.21 & 58.42
& 58.17 & 66.1 \\
VScan~\citep{vscan2025} & 50\%
& 72.32 & \textbf{78.51} & 64.95
& 66.70 & \textbf{74.32} & 55.52
& 67.31 & 66.72
& 68.29 & 77.5 \\
\midrule
\multicolumn{12}{@{}l}{\textit{Adaptive: 10\% $\rightarrow$ 100\% tokens}} \\
\rowcolor{lightblue}
\textbf{SmartRes-Pro} & 42\%
& \textbf{74.19} & 77.87 & \textbf{70.34}
& \textbf{67.68} & 71.31 & \textbf{61.97}
& \textbf{71.72} & \textbf{71.98}
& \textbf{70.88} & \textbf{80.5} \\
\bottomrule
\end{tabularx}
\vspace{-4mm}
\end{table*}
\vspace{-2mm}
\subsection{Evaluation on Small Objects}
\label{subsec:small}
\vspace{-1mm}
Small objects are common in egocentric grounding and are especially sensitive to resolution loss.
We therefore analyze performance by object scale in Fig.~\ref{fig:results}\,(b).
The results show that performance degradation under compression is scale-dependent.
Uniform down-scaling to 50\% reduces $\mathrm{P}_{\mathrm{s}}$ by 10.7\% on Ego4D and 7.8\% on EgoIntention, while the corresponding drops for $\mathrm{P}_{\mathrm{l}}$ are only 6.3\% and 5.8\%, respectively.
This indicates that medium and large objects remain well-represented after down-sampling, whereas small objects are more sensitive to resolution loss.
Token pruning methods further amplify this imbalance.
In contrast, \emph{SmartRes-Pro} better preserves small-object performance under a reduced token budget.
At a 55\% ratio, it retains {88.9\%} of the full-resolution $\mathrm{P}_{\mathrm{s}}$ on Ego4D and {89.5\%} on EgoIntention, while maintaining competitive performance on medium and large objects.
These results show that routing high-resolution computation to grounding-relevant regions mitigates small-object resolution loss in egocentric scenes.
\vspace{-2mm}
\subsection{Generalization to Standard REC Benchmarks}
\vspace{-1mm}
\label{subsec:rec}
We further evaluate SmartRes on standard referring expression comprehension (REC) benchmarks in Tab.~\ref{tab:refcoco_main}. SmartRes generalizes well to general visual grounding.
Averaged over the eight evaluation splits, SmartRes-Pro achieves 70.88 Overall P$@0.5$ and 80.5\% performance retention while using only 42\% of tokens on average.
Our results also show that uniform down-scaling provides a competitive low-cost baseline on REC.
At a 50\% token ratio, it achieves 64.78 Overall P$@0.5$, which drops to 54.11 and 37.72 at 32\% and 10\%, respectively.
\emph{SmartRes-Pro} improves over Down-scaling-50\% and VScan by 6.10 and 2.59 Overall P$@0.5$, respectively.
\vspace{-2mm}
\subsection{Qualitative Analysis}
\begin{figure*}[t!]
  \centering
  \includegraphics[width=\linewidth]{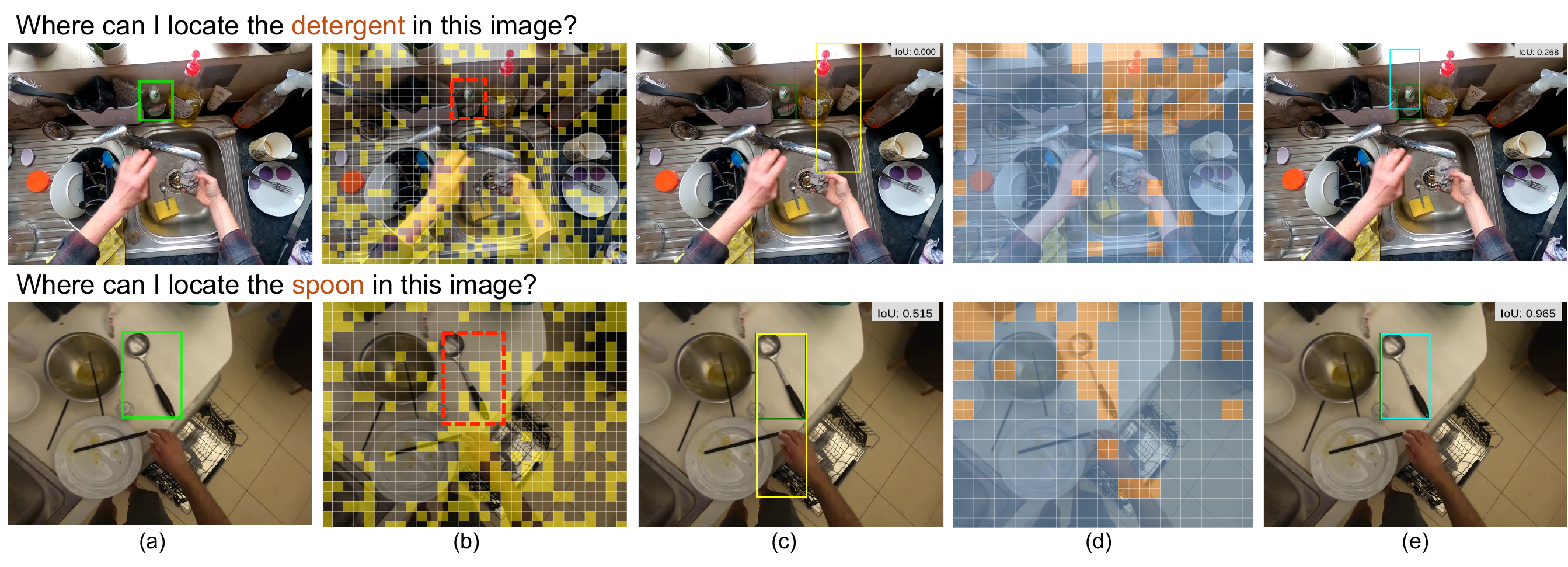}
  \vspace{-0.5cm}
  \caption{
  \textbf{Visualization of compared efficiency mechanisms.}
  (a) Input image with query.
  \textbf{FastV's} (b) pruning mask (\textcolor[RGB]{255,230,153}{\rule[-0.1ex]{1.5ex}{1.5ex}}) and (c) prediction. Relying on noisy latent proxies, the baseline often discards fine-grained details, missing small targets like the ``detergent'' in Row~1.
  \textbf{SmartRes'} (d) routing mask and (e) prediction.
  By dynamically routing high-resolution computation (\textcolor[RGB]{244,177,131}{\rule[-0.1ex]{1.5ex}{1.5ex}}) to object-centric areas against a low-res background (\textcolor[RGB]{189,215,238}{\rule[-0.1ex]{1.5ex}{1.5ex}}), our method yields precise localization.
  \label{fig:visualizations}
  \vspace{-2mm}
  }
\end{figure*}

\vspace{-1mm}
Fig.~\ref{fig:visualizations} shows that FastV discards fine‑grained details, causing it to miss small targets like ``detergent'' in row~1. In contrast, SmartRes routes high‑resolution computation to object‑centric regions with a low‑resolution context. Similarly, the token retention map derived from FastV (Fig.~\ref{fig:visualizations}(b)) is divergent, which ignores ``spoon'', while ours focuses on object-aware regions. In addition to the accurate localization results, this further enhances the interpretability of \emph{SmartRes}.
\begin{table*}[ht]
\centering
\scriptsize
\caption{
Ablation of key components on EgoIntention.
Overall denotes the mean P$@0.5$ over the two EgoIntention splits.
Let $\mathcal{G}$ be the ground-truth foreground token set and $\mathcal{R}$ be the set of tokens routed to HR refinement.
FG-Recall denotes $|\mathcal{R}\cap\mathcal{G}|/|\mathcal{G}|$, measuring foreground coverage, while FG-Precision denotes $|\mathcal{R}\cap\mathcal{G}|/|\mathcal{R}|$.
}
\vspace{-2mm}
\setlength{\tabcolsep}{3.0pt}
\renewcommand{\arraystretch}{1.05}
\begin{subtable}[t]{0.54\textwidth}
\centering
\begin{tabularx}{\linewidth}{@{}X c c c c@{}}
\toprule
\textbf{Method} & \textbf{Ratio} & \textbf{Overall} & \textbf{FG-Recall} & \textbf{FG-Precision} \\
& & P$@0.5$ & (\%) & (\%) \\
\midrule
Uniform Down-sampling & 32.14\% & 27.47 & -- & -- \\
SmartRes-Random & 33.00\% & 32.15 & 25.6 & 19.1 \\
Attn2HR & 32.96\% & 36.12 & 18.5 & 22.3 \\
\midrule
SmartRes w/o $\mathcal{L}_{\mathrm{hinge}}$ & 32.94\% & 45.35 & 37.0 & 24.7 \\
\rowcolor{gray!10}
\textbf{SmartRes (Full)} & 33.05\% & \textbf{49.39} & {54.0} & {31.1} \\
\midrule
\textit{Oracle (GT-guided)} & 18.97\% & 58.45 & 100.0 & 100.0 \\
\bottomrule
\end{tabularx}
\caption{Routing strategy and supervision.}
\label{tab:ablation_strategy}
\end{subtable}
\hfill
\begin{subtable}[t]{0.43\textwidth}
\centering
\begin{tabular}{@{}l c c c c@{}}
\toprule
\textbf{Setting} & \textbf{Value} & \textbf{Overall} & \textbf{FG-Recall} & \textbf{FLOPs} \\
& & P$@0.5$ & (\%) & (T) \\
\midrule
\multirow{3}{*}{Layer $\ell$}
& 2nd  & 47.82 & 47.1 & 3.82 \\
& 15th & 48.95 & 51.8 & 4.01 \\
& \cellcolor{gray!10}30th
& \cellcolor{gray!10}49.39
& \cellcolor{gray!10}54.2
& \cellcolor{gray!10}4.05 \\
\midrule
\multirow{3}{*}{Threshold $\tau$}
& 0.3 & \textbf{49.41} & \textbf{55.6} & 4.82 \\
& \cellcolor{gray!10}0.5
& \cellcolor{gray!10}49.39
& \cellcolor{gray!10}54.2
& \cellcolor{gray!10}4.05 \\
& 0.7 & 46.20 & 42.5 & \textbf{3.10} \\
\bottomrule
\end{tabular}
\caption{Routing layer and threshold.}
\label{tab:ablation_routing}
\end{subtable}
\label{tab:ablation}
\vspace{-6mm}
\end{table*}
\vspace{-2mm}
\subsection{Ablation Studies}
\label{subsec:ablation}
\vspace{-1mm}
\noindent\textbf{Component analysis.}
Tab.~\ref{tab:ablation_strategy} ablates the main components of SmartRes.
Compared with uniform down-sampling, random HR patch selection improves Overall P$@0.5$ by 4.68\%, showing that preserving local high-resolution details is already beneficial.
Using the learned router raises Overall P$@0.5$ to 45.35, and adding the margin-regularized loss $\mathcal{L}_{\mathrm{hinge}}$ further improves it to 49.39.
This objective also boosts FG-Recall from 37.0\% to 54.0\% and FG-Precision from 24.7\% to 31.1\%, effectively mitigating foreground--background imbalance.
The result of Oracle (GT-guided) (58.45) indicates remaining potential for precise patch selection.
  
\noindent\textbf{Attn2HR.}
We compare against \textit{Attn2HR}, a fixed attention‑guided selection baseline detailed in Sec.~\ref{sec:attn2hr_crop} of the \textit{Supplementary}.
The results in Tab.~\ref{tab:ablation_strategy} show that it improves over uniform down‑sampling and random routing, raising Overall P$@0.5$ from 32.15\% to 36.12\%.
Thus, visual self‑attention provides useful saliency cues for high‑resolution selection.
However, it remains substantially below the learned router under a comparable token budget, indicating that fixed attention heuristics are insufficient for guiding patch selection and highlighting the need for learnable patch‑level routing.
  
\noindent\textbf{Router hyperparameters.}
Tab.~\ref{tab:ablation_routing} studies the routing layer $\ell$ and threshold $\tau$. Using deeper features improves routing quality. Increasing $\ell$ from 2nd to 30th raises Overall P$@$0.5 from 47.82 to 49.39 and FG-Recall from 47.1\% to 54.2\%. Besides, the smaller $\tau$=$0.3$ activates more object-aware regions, thus increasing the performance of Overall P$@$0.5 from $49.39$ to $49.41$ at the cost of efficiency. Conversely, a variant with a larger $\tau$=0.7 improves efficiency while degrading performance.
\vspace{-3mm}

\vspace{-2mm}
\section{Conclusion}
\vspace{-3mm}
Egocentric visual grounding requires high-resolution input to perceive fine-grained details, yet processing such frames is computationally expensive. To address this, we introduce SmartRes, a framework that performs dynamic resolution routing in the pixel space. Using a low-resolution branch as guidance, it selectively activates high-resolution patches only in object regions, preserving fine details while lowering computation. Evaluations demonstrate that SmartRes maintains high accuracy with significantly fewer tokens, offering a practical and efficient solution for deploying grounding models on edge devices. Future work can extend SmartRes to egocentric videos.

\bibliographystyle{unsrtnat}
\bibliography{main}

@String(CVPR= {IEEE Conf. Comput. Vis. Pattern Recog.})

@String(ICCV= {Int. Conf. Comput. Vis.})

@String(ICLR = {Int. Conf. Learn. Represent.})

@String(CVPR  = {CVPR})

@String(ICCV  = {ICCV})

@String(ICLR  = {ICLR})

@article{zhang2025mllms,
  title={Mllms know where to look: Training-free perception of small visual details with multimodal llms},
  author={Zhang, Jiarui and Khayatkhoei, Mahyar and Chhikara, Prateek and Ilievski, Filip},
  journal=ICLR,
  year={2025}
}

@article{bai2023qwen,
  title={Qwen-vl: A versatile vision-language model for understanding, localization, text reading, and beyond},
  author={Bai, Jinze and Bai, Shuai and Yang, Shusheng and Wang, Shijie and Tan, Sinan and Wang, Peng and Lin, Junyang and Zhou, Chang and Zhou, Jingren},
  journal={arXiv preprint arXiv:2308.12966},
  volume={1},
  number={2},
  pages={3},
  year={2023}
}

@article{bai2025qwen2,
  title={{Qwen2.5-VL} technical report},
  author={Bai, Shuai and Chen, Keqin and Liu, Xuejing and Wang, Jialin and Ge, Wenbin and Song, Sibo and Dang, Kai and Wang, Peng and Wang, Shijie and Tang, Jun and others},
  journal={arXiv preprint arXiv:2502.13923},
  year={2025}
}

@article{wang2024qwen2,
  title={{Qwen2-VL}: Enhancing Vision-Language Model's Perception of the World at Any Resolution},
  author={Wang, Peng and Bai, Shuai and Tan, Sinan and Wang, Shijie and Fan, Zhihao and Bai, Jinze and Chen, Keqin and Liu, Xuejing and Wang, Jialin and Ge, Wenbin and others},
  journal={arXiv preprint arXiv:2409.12191},
  year={2024}
}

@inproceedings{zhang2024llava,
  title={Llava-grounding: Grounded visual chat with large multimodal models},
  author={Zhang, Hao and Li, Hongyang and Li, Feng and Ren, Tianhe and Zou, Xueyan and Liu, Shilong and Huang, Shijia and Gao, Jianfeng and Zhang, Lei and Li, Chunyuan and others},
  booktitle={European Conference on Computer Vision},
  pages={19--35},
  year={2024},
  organization={Springer}
}

@misc{liu2023llava,
      title={Visual Instruction Tuning}, 
      author={Liu, Haotian and Li, Chunyuan and Wu, Qingyang and Lee, Yong Jae},
      publisher={NeurIPS},
      year={2023},
}

@article{sun2025visual,
  title={Visual Intention Grounding for Egocentric Assistants},
  author={Sun, Pengzhan and Xiao, Junbin and Tse, Tze Ho Elden and Li, Yicong and Akula, Arjun and Yao, Angela},
  journal=ICCV,
  year={2025}
}

@inproceedings{kazemzadeh2014referitgame,
  title={{ReferItGame}: Referring to objects in photographs of natural scenes},
  author={Kazemzadeh, Sahar and Ordonez, Vicente and Matten, Mark and Berg, Tamara},
  booktitle={Proceedings of the 2014 conference on empirical methods in natural language processing (EMNLP)},
  pages={787--798},
  year={2014}
}

@inproceedings{mao2016generation,
  title={Generation and comprehension of unambiguous object descriptions},
  author={Mao, Junhua and Huang, Jonathan and Toshev, Alexander and Camburu, Oana and Yuille, Alan L and Murphy, Kevin},
  booktitle={Proceedings of the IEEE conference on computer vision and pattern recognition},
  pages={11--20},
  year={2016}
}

@inproceedings{nagaraja2016modeling,
  title={Modeling context between objects for referring expression understanding},
  author={Nagaraja, Varun K and Morariu, Vlad I and Davis, Larry S},
  booktitle={Computer Vision--ECCV 2016: 14th European Conference, Amsterdam, The Netherlands, October 11--14, 2016, Proceedings, Part IV 14},
  pages={792--807},
  year={2016},
  organization={Springer}
}

@inproceedings{yu2016modeling,
  title={Modeling context in referring expressions},
  author={Yu, Licheng and Poirson, Patrick and Yang, Shan and Berg, Alexander C and Berg, Tamara L},
  booktitle={Computer Vision--ECCV 2016: 14th European Conference, Amsterdam, The Netherlands, October 11-14, 2016, Proceedings, Part II 14},
  pages={69--85},
  year={2016},
  organization={Springer}
}

@inproceedings{grauman2022ego4d,
  title={Ego4d: Around the world in 3,000 hours of egocentric video},
  author={Grauman, Kristen and Westbury, Andrew and Byrne, Eugene and Chavis, Zachary and Furnari, Antonino and Girdhar, Rohit and Hamburger, Jackson and Jiang, Hao and Liu, Miao and Liu, Xingyu and others},
  booktitle={Proceedings of the IEEE/CVF Conference on Computer Vision and Pattern Recognition},
  pages={18995--19012},
  year={2022}
}

@inproceedings{ramanathan2023paco,
  title={Paco: Parts and attributes of common objects},
  author={Ramanathan, Vignesh and Kalia, Anmol and Petrovic, Vladan and Wen, Yi and Zheng, Baixue and Guo, Baishan and Wang, Rui and Marquez, Aaron and Kovvuri, Rama and Kadian, Abhishek and others},
  booktitle={Proceedings of the IEEE/CVF Conference on Computer Vision and Pattern Recognition},
  pages={7141--7151},
  year={2023}
}

@article{damen2022rescaling,
   title={Rescaling Egocentric Vision},
   author={Damen, Dima and Doughty, Hazel and Farinella, Giovanni Maria and Furnari, Antonino and Ma, Jian and Kazakos, Evangelos and Moltisanti, Davide and Munro, Jonathan and Perrett, Toby and Price, Will and Wray, Michael},
           journal={International Journal of Computer Vision},
           volume={130},
           number={1},
           pages={33--55},
           year={2022},
           publisher={Springer}
}

@inproceedings{mitchell2013generating,
  title={Generating expressions that refer to visible objects},
  author={Mitchell, Margaret and Van Deemter, Kees and Reiter, Ehud},
  booktitle={Proceedings of the 2013 Conference of the North American Chapter of the Association for Computational Linguistics: Human Language Technologies},
  pages={1174--1184},
  year={2013}
}

@inproceedings{fitzgerald2013learning,
  title={Learning distributions over logical forms for referring expression generation},
  author={FitzGerald, Nicholas and Artzi, Yoav and Zettlemoyer, Luke},
  booktitle={Proceedings of the 2013 conference on empirical methods in natural language processing},
  pages={1914--1925},
  year={2013}
}

@article{liu2024llavanext,
  title   = {{LLaVA-NeXT}: Improved Reasoning, OCR, and World Knowledge},
  author  = {Liu, Haotian and Li, Chunyuan and Li, Yuheng and Li, Bo and Zhang, Yuanhan and Shen, Sheng and Lee, Yong Jae},
  journal = {arXiv preprint},
  year    = {2024}
}

@article{bolya2022token,
  title={Token merging: Your vit but faster},
  author={Bolya, Daniel and Fu, Cheng-Yang and Dai, Xiaoliang and Zhang, Peizhao and Feichtenhofer, Christoph and Hoffman, Judy},
  journal={arXiv preprint arXiv:2210.09461},
  year={2022}
}

@article{zhang2024sparsevlm,
  title={SparseVLM: Visual Token Sparsification for Efficient Vision-Language Model Inference},
  author={Zhang, Yuan and Fan, Chun-Kai and Ma, Junpeng and Zheng, Wenzhao and Huang, Tao and Cheng, Kuan and Gudovskiy, Denis and Okuno, Tomoyuki and Nakata, Yohei and Keutzer, Kurt and others},
  journal={arXiv preprint arXiv:2410.04417},
  year={2024}
}

@article{xing2024pyramiddrop,
  title={PyramidDrop: Accelerating your large vision-language models via pyramid visual redundancy reduction},
  author={Xing, Long and Huang, Qidong and Dong, Xiaoyi and Lu, Jiajie and Zhang, Pan and Zang, Yuhang and Cao, Yuhang and He, Conghui and Wang, Jiaqi and Wu, Feng and others},
  journal={arXiv preprint arXiv:2410.17247},
  year={2024}
}

@inproceedings{yang2025visionzip,
  title={{VisionZip}: Longer is better but not necessary in vision language models},
  author={Yang, Senqiao and Chen, Yukang and Tian, Zhuotao and Wang, Chengyao and Li, Jingyao and Yu, Bei and Jia, Jiaya},
  booktitle={Proceedings of the Computer Vision and Pattern Recognition Conference},
  pages={19792--19802},
  year={2025}
}

@article{darcet2023vision,
  title={Vision transformers need registers},
  author={Darcet, Timoth{\'e}e and Oquab, Maxime and Mairal, Julien and Bojanowski, Piotr},
  journal={arXiv preprint arXiv:2309.16588},
  year={2023}
}

@inproceedings{sun2019deep,
  title={Deep high-resolution representation learning for human pose estimation},
  author={Sun, Ke and Xiao, Bin and Liu, Dong and Wang, Jingdong},
  booktitle={Proceedings of the IEEE/CVF conference on computer vision and pattern recognition},
  pages={5693--5703},
  year={2019}
}

@article{yao2026towards,
  title={Towards Efficient Multimodal Large Language Models: A Survey on Token Compression},
  author={Yao, Linli and Xing, Long and Shi, Yang and Li, Sida and Liu, Yuanxin and Dong, Yuhao and Zhang, Yi-Fan and Li, Lei and Dong, Qingxiu and Dong, Xiaoyi and others},
  journal={Authorea Preprints},
  year={2026},
  publisher={Authorea}
}

@article{zou2025don,
  title={Don't Just Chase ``Highlighted Tokens'' in {MLLM}s: Revisiting Visual Holistic Context Retention},
  author={Zou, Xin and Lu, Di and Wang, Yizhou and Yan, Yibo and Lyu, Yuanhuiyi and Zheng, Xu and Zhang, Linfeng and Hu, Xuming},
  journal={arXiv preprint arXiv:2510.02912},
  year={2025}
}

@article{li2025tokenpacker,
  title={{TokenPacker}: Efficient visual projector for multimodal llm},
  author={Li, Wentong and Yuan, Yuqian and Liu, Jian and Tang, Dongqi and Wang, Song and Qin, Jie and Zhu, Jianke and Zhang, Lei},
  journal={International Journal of Computer Vision},
  pages={1--19},
  year={2025},
  publisher={Springer}
}

@inproceedings{huang2025hires,
  title={{HiRes-LLaVA}: Restoring fragmentation input in high-resolution large vision-language models},
  author={Huang, Runhui and Ding, Xinpeng and Wang, Chunwei and Han, Jianhua and Liu, Yulong and Zhao, Hengshuang and Xu, Hang and Hou, Lu and Zhang, Wei and Liang, Xiaodan},
  booktitle={Proceedings of the Computer Vision and Pattern Recognition Conference},
  pages={29814--29824},
  year={2025}
}

@article{bengio2013estimating,
  title={Estimating or propagating gradients through stochastic neurons for conditional computation},
  author={Bengio, Yoshua and L{\'e}onard, Nicholas and Courville, Aaron},
  journal={arXiv preprint arXiv:1308.3432},
  year={2013}
}

@inproceedings{chen2024image,
  title={An image is worth 1/2 tokens after layer 2: Plug-and-play inference acceleration for large vision-language models},
  author={Chen, Liang and Zhao, Haozhe and Liu, Tianyu and Bai, Shuai and Lin, Junyang and Zhou, Chang and Chang, Baobao},
  booktitle={European Conference on Computer Vision},
  pages={19--35},
  year={2024},
  organization={Springer}
}

@inproceedings{kurita2023refego,
  title={{RefEgo}: Referring expression comprehension dataset from first-person perception of ego4d},
  author={Kurita, Shuhei and Katsura, Naoki and Onami, Eri},
  booktitle={Proceedings of the IEEE/CVF International Conference on Computer Vision},
  pages={15214--15224},
  year={2023}
}

@inproceedings{ye2025atpllava,
  title     = {{ATP-LLaVA}: Adaptive Token Pruning for Large Vision Language Models},
  author    = {Ye, Xubing and Gan, Yukang and Ge, Yixiao and Zhang, Xiao-Ping and Tang, Yansong},
  booktitle = {Proceedings of the IEEE/CVF Conference on Computer Vision and Pattern Recognition},
  year      = {2025}
}

@inproceedings{jiang2025teva,
  title     = {Token-Efficient VLM: High-Resolution Image Understanding via Dynamic Region Proposal},
  author    = {Jiang, Yitong and Gu, Jinwei and Xue, Tianfan and Cheung, Ka Chun and Molchanov, Pavlo and Yin, Hongxu and Liu, Sifei},
  booktitle = {Proceedings of the IEEE/CVF International Conference on Computer Vision},
  pages     = {24147--24158},
  year      = {2025}
}

@inproceedings{lin2017focal,
  title     = {Focal Loss for Dense Object Detection},
  author    = {Lin, Tsung-Yi and Goyal, Priya and Girshick, Ross B. and He, Kaiming and Doll{\'a}r, Piotr},
  booktitle = {Proceedings of the IEEE International Conference on Computer Vision (ICCV)},
  pages     = {2980--2988},
  year      = {2017},
  doi       = {10.1109/ICCV.2017.324}
}

@inproceedings{milletari2016vnet,
  title     = {{V-Net}: Fully Convolutional Neural Networks for Volumetric Medical Image Segmentation},
  author    = {Milletari, Fausto and Navab, Nassir and Ahmadi, Seyed-Ahmad},
  booktitle = {2016 Fourth International Conference on 3D Vision (3DV)},
  pages     = {565--571},
  year      = {2016},
  doi       = {10.1109/3DV.2016.79},
  publisher = {IEEE}
}

@inproceedings{salehi2017tversky,
  title     = {Tversky Loss Function for Image Segmentation Using 3D Fully Convolutional Deep Networks},
  author    = {Salehi, Seyed Sadegh Mohseni and Erdogmus, Deniz and Gholipour, Ali},
  booktitle = {Machine Learning in Medical Imaging},
  series    = {Lecture Notes in Computer Science},
  volume    = {10541},
  pages     = {379--387},
  year      = {2017},
  editor    = {Wang, Qian and Shi, Yinghuan and Suk, Heung-Il and Suzuki, Kenji},
  publisher = {Springer},
  address   = {Cham},
  doi       = {10.1007/978-3-319-67389-9_44}
}

@article{huang2024dynamic,
  title={Dynamic-{LLaVA}: Efficient multimodal large language models via dynamic vision-language context sparsification},
  author={Huang, Wenxuan and Zhai, Zijie and Shen, Yunhang and Cao, Shaosheng and Zhao, Fei and Xu, Xiangfeng and Ye, Zheyu and Hu, Yao and Lin, Shaohui},
  journal={arXiv preprint arXiv:2412.00876},
  year={2024}
}

@inproceedings{shi2025ps3,
  title     = {Scaling Vision Pre-Training to {4K} Resolution},
  author    = {Shi, Baifeng and Li, Boyi and Cai, Han and Lu, Yao and Liu, Sifei and Pavone, Marco and Kautz, Jan and Han, Song and Darrell, Trevor and Molchanov, Pavlo and Yin, Hongxu},
  booktitle = {Proceedings of the IEEE/CVF Conference on Computer Vision and Pattern Recognition (CVPR)},
  pages     = {9631--9640},
  month     = {June},
  year      = {2025},
  eprint    = {2503.19903},
  archivePrefix = {arXiv},
  primaryClass  = {cs.CV}
}

@inproceedings{yang2025visionthink,
  title     = {{VisionThink}: Smart and Efficient Vision Language Model via Reinforcement Learning},
  author    = {Yang, Senqiao and Li, Junyi and Lai, Xin and Wu, Jinming and Li, Wei and Ma, Zejun and Yu, Bei and Zhao, Hengshuang and Jia, Jiaya},
  booktitle = {Advances in Neural Information Processing Systems},
  volume    = {38},
  year      = {2025},
  eprint    = {2507.13348},
  archivePrefix = {arXiv},
  primaryClass  = {cs.CV}
}

@article{vscan2025,
  title   = {{VScan}: Rethinking Visual Token Reduction for Efficient Large Vision-Language Models},
  author  = {Zhang, Ce and Ma, Kaixin and Fang, Tianqing and Yu, Wenhao and Zhang, Hongming and Zhang, Zhisong and Mi, Haitao and Yu, Dong},
  journal = {Transactions on Machine Learning Research},
  year    = {2026},
  eprint  = {2505.22654},
  archivePrefix = {arXiv},
  primaryClass  = {cs.CV}
}

@article{bai2025altp,
  title   = {Local Information Matters: Inference Acceleration for Grounded Conversation Generation Models Through Adaptive Local-Aware Token Pruning},
  author  = {Bai, Bizhe and Cao, Jianjian and Luo, Yadan and Chen, Tao},
  journal = {arXiv preprint arXiv:2503.23959},
  year    = {2025},
  doi     = {10.48550/arXiv.2503.23959},
  eprint  = {2503.23959},
  archivePrefix = {arXiv},
  primaryClass  = {cs.CV}
}

@inproceedings{focusui,
  title     = {{FocusUI}: Efficient {UI} Grounding via Position-Preserving Visual Token Selection},
  author    = {Ouyang, Mingyu and Lin, Kevin Qinghong and Shou, Mike Zheng and Ng, Hwee Tou},
  booktitle = {Proceedings of the IEEE/CVF Conference on Computer Vision and Pattern Recognition (CVPR)},
  year      = {2026},
  eprint    = {2601.03928},
  archivePrefix = {arXiv},
  primaryClass  = {cs.CV}
}

@inproceedings{wu2024vstar,
  title={{V*}: Guided Visual Search as a Core Mechanism in Multimodal {LLMs}},
  author={Wu, Penghao and Xie, Saining},
  booktitle={Proceedings of the IEEE/CVF Conference on Computer Vision and Pattern Recognition (CVPR)},
  pages={13084--13094},
  year={2024}
}

@article{qian2025zoomer,
  title   = {Zoomer: Adaptive Image Focus Optimization for Black-box MLLM},
  author  = {Qian, Jiaxu and Wang, Chendong and Yang, Yifan and Zhang, Chaoyun and Jiang, Huiqiang and Luo, Xufang and Kang, Yu and Lin, Qingwei and Zhang, Anlan and Jiang, Shiqi and Cao, Ting and Mao, Tianjun and Banerjee, Suman and Liu, Guyue and Rajmohan, Saravan and Zhang, Dongmei and Yang, Yuqing and Zhang, Qi and Qiu, Lili},
  journal = {arXiv preprint arXiv:2505.00742},
  year    = {2025}
}

@article{lin2025adaptvision,
  title   = {{AdaptVision}: Efficient Vision-Language Models via Adaptive Visual Acquisition},
  author  = {Lin, Zichuan and Liu, Yicheng and Yang, Yang and Tao, Lvfang and Ye, Deheng},
  journal = {arXiv preprint arXiv:2512.03794},
  year    = {2025}
}

@inproceedings{su2026padt,
  title     = {Patch-as-Decodable-Token: Towards Unified Multi-Modal Vision Tasks in MLLMs},
  author    = {Su, Yongyi and Zhang, Haojie and Li, Shijie and Liu, Nanqing and Liao, Jingyi and Pan, Junyi and Liu, Yuan and Xing, Xiaofen and Sun, Chong and Li, Chen and Chen, Nancy F. and Yan, Shuicheng and Yang, Xulei and Xu, Xun},
  booktitle = {International Conference on Learning Representations (ICLR)},
  year      = {2026}
}

@article{bai2025qwen3,
  title={{Qwen3-VL} technical report},
  author={Bai, Shuai and Cai, Yuxuan and Chen, Ruizhe and Chen, Keqin and Chen, Xionghui and Cheng, Zesen and Deng, Lianghao and Ding, Wei and Gao, Chang and Ge, Chunjiang and others},
  journal={arXiv preprint arXiv:2511.21631},
  year={2025}
}

@article{he2009learning,
  title={Learning from Imbalanced Data},
  author={He, Haibo and Garcia, Edwardo A.},
  journal={IEEE Transactions on Knowledge and Data Engineering},
  volume={21},
  number={9},
  pages={1263--1284},
  year={2009}
}

@inproceedings{cui2019classbalanced,
  title={Class-Balanced Loss Based on Effective Number of Samples},
  author={Cui, Yin and Jia, Menglin and Lin, Tsung-Yi and Song, Yang and Belongie, Serge},
  booktitle={Proceedings of the IEEE/CVF Conference on Computer Vision and Pattern Recognition},
  pages={9268--9277},
  year={2019}
}

\newpage
\appendix
\onecolumn
\section{Image Preparation and Token Correspondence}
\label{supp:processing}
Given an input image $\mathbf I\in\mathbb R^{H\times W\times 3}$ and target token ratios
$r_{\mathrm{LR}}$ and $r_{\mathrm{HR}}$, we first compute the target spatial resolution for each branch.
For branch $b\in\{\mathrm{LR},\mathrm{HR}\}$ with ratio $r_b$, we set
\begin{equation}
\widetilde H_b = H\sqrt{r_b},
\qquad
\widetilde W_b = W\sqrt{r_b}.
\end{equation}
The isotropic scaling preserves the original aspect ratio while approximately matching the desired token ratio.
To satisfy the patching and spatial-merging constraints of the vision encoder, we snap each dimension to a multiple of the effective visual-token stride $F=PM$:
\begin{equation}
H'_b=\operatorname{snap}_F(\widetilde H_b),
\qquad
W'_b=\operatorname{snap}_F(\widetilde W_b),
\qquad
b\in\{\mathrm{LR},\mathrm{HR}\}.
\end{equation}
Here,
\begin{equation}
\operatorname{snap}_F(s)
=
F\cdot
\max\left(1,\left\lfloor \frac{s}{F}+\frac{1}{2}\right\rfloor\right)
\label{eq:snap}
\end{equation}
rounds $s$ to the nearest positive multiple of $F$.
For the Qwen2.5-VL~\cite{bai2025qwen2} example, the patch size is $P=14$ and the spatial merge size is $M=2$, so $F=28$.
We resize $\mathbf I$ to $(H'_{\mathrm{LR}},W'_{\mathrm{LR}})$ and
$(H'_{\mathrm{HR}},W'_{\mathrm{HR}})$ to obtain
$\mathbf I_{\mathrm{LR}}$ and $\mathbf I_{\mathrm{HR}}$, respectively.
The snapped dimensions induce routing grids:
\begin{equation}
G^h_b=\frac{H'_b}{F},
\qquad
G^w_b=\frac{W'_b}{F},
\qquad
N_b=G^h_bG^w_b,
\end{equation}
where $G^h_b$ and $G^w_b$ denote the height and width of the routing grid for branch $b$, and $N_b$ is the corresponding feature length used by the router.
Because of snapping, the realized token ratio may differ slightly from the target ratio.
The LR and HR routing grids are aligned by a deterministic coordinate mapping.
Each HR location at grid coordinate $(y_{\mathrm{HR}},x_{\mathrm{HR}})$ is assigned to a unique LR parent:
\begin{equation}
\label{eq:hr2lr_snap}
\phi(y_{\mathrm{HR}},x_{\mathrm{HR}})
=
\left(
\left\lfloor
\frac{y_{\mathrm{HR}}G^h_{\mathrm{LR}}}{G^h_{\mathrm{HR}}}
\right\rfloor,\;
\left\lfloor
\frac{x_{\mathrm{HR}}G^w_{\mathrm{LR}}}{G^w_{\mathrm{HR}}}
\right\rfloor
\right).
\end{equation}
Here,
\begin{equation}
\label{eq:hr_lr_domain}
\phi:
\{0,\ldots,G^h_{\mathrm{HR}}-1\}
\times
\{0,\ldots,G^w_{\mathrm{HR}}-1\}
\rightarrow
\{0,\ldots,G^h_{\mathrm{LR}}-1\}
\times
\{0,\ldots,G^w_{\mathrm{LR}}-1\}.
\end{equation}
This mapping provides the HR-to-LR correspondence used to upsample the routing mask and select HR patches.
\section{Dataset Statistics}
\label{sec:dataset_statistics}
\begin{table}[h]
\centering
\small
\setlength{\tabcolsep}{14pt}
\renewcommand{\arraystretch}{1.08}
\caption{\textbf{Dataset statistics for egocentric grounding and standard REC benchmarks.}
RefCOCO family denotes the combined statistics of RefCOCO, RefCOCO+, and RefCOCOg.
Avg. Res. reports the mean image resolution.}
\label{tab:dataset_stats}
\begin{tabular}{@{}lccc@{}}
\toprule
\multirow{2}{*}{\textbf{Metric}}
& \multicolumn{2}{c}{\textbf{Egocentric Grounding}}
& \textbf{Standard REC} \\
\cmidrule(lr){2-3} \cmidrule(lr){4-4}
& \textbf{Ego4D} & \textbf{EgoIntention} & \textbf{RefCOCO family} \\
\midrule
\# Images  & 26.4K & 35.6K & 28.2K \\
\# Boxes   & 58.4K & 51.1K & 321.3K \\
Avg. Res.  & $1825 \times 1270$ & $1844 \times 1297$ & $554 \times 486$ \\
\midrule
\rowcolor{gray!10}
\multicolumn{4}{@{}l}{\textit{Object scale distribution}} \\
Small  & 17.07\% & 10.07\% & 0.00\% \\
Medium & 54.95\% & 44.92\% & 3.31\% \\
Large  & 27.99\% & 45.01\% & 96.69\% \\
\bottomrule
\end{tabular}
\end{table}
Tab.~\ref{tab:dataset_stats} highlights the resolution and object-scale differences between egocentric grounding benchmarks and standard referring expression comprehension (REC) datasets.
Egocentric datasets have substantially higher native resolutions, with average image areas are more than $8\times$ larger than those of the RefCOCO family.
They also exhibit a markedly different object-scale distribution.
Small and medium objects account for 72.02\% of instances in Ego4D and 54.99\% in EgoIntention, whereas the RefCOCO family is  dominated by large objects.
These statistics motivate resolution-aware processing for egocentric grounding, where many targets occupy limited spatial extent despite high-resolution inputs.
\section{Gradient Analysis of Imbalance-Aware Routing Losses}
\label{app:imbalance_losses}
We compare the proposed margin hinge regularizer in Eq.~\eqref{eq:Lhinge} with  imbalance-aware objectives for binary routing: vanilla BCE, focal loss~\citep{lin2017focal}, and soft Dice/Tversky losses~\citep{milletari2016vnet,salehi2017tversky}.
Let $N=N_{\mathrm{LR}}$, $z_i$ be the routing logit for LR token $i$, $p_i=\sigma(z_i)$, and $\hat M_i\in\{0,1\}$ be the routing label.
We denote the foreground and background token sets by
$\mathcal{P}=\{i:\hat M_i=1\}$ and $\mathcal{N}=\{i:\hat M_i=0\}$, with $|\mathcal{P}|\ll|\mathcal{N}|$ in egocentric grounding.
For each loss, we inspect the logit gradient $\partial\mathcal{L}/\partial z_i$ and the class-level gradient masses
\begin{equation}
G_{\mathcal{P}}=\sum_{i\in\mathcal{P}}
\left|\frac{\partial\mathcal{L}}{\partial z_i}\right|,
\qquad
G_{\mathcal{N}}=\sum_{i\in\mathcal{N}}
\left|\frac{\partial\mathcal{L}}{\partial z_i}\right|.
\end{equation}
These quantities measure the aggregate update pressure applied to foreground and background tokens.

\noindent\textbf{BCE.}
The vanilla BCE loss is
\begin{equation}
\mathcal{L}_{\mathrm{BCE}}
=
-\frac{1}{N}
\sum_i
\left[
\hat M_i\log p_i+
(1-\hat M_i)\log(1-p_i)
\right],
\end{equation}
with gradient
\begin{equation}
\frac{\partial \mathcal{L}_{\mathrm{BCE}}}{\partial z_i}
=
\frac{1}{N}(p_i-\hat M_i).
\end{equation}
Thus,
\begin{equation}
G_{\mathcal{P}}
=
\frac{1}{N}\sum_{i\in\mathcal{P}}(1-p_i),
\qquad
G_{\mathcal{N}}
=
\frac{1}{N}\sum_{i\in\mathcal{N}}p_i .
\end{equation}
BCE has no class-level normalization.
When the average foreground and background errors are comparable, the ratio $G_{\mathcal{P}}/G_{\mathcal{N}}$ scales with $|\mathcal{P}|/|\mathcal{N}|$, so the aggregate update can be dominated by the background tokens.

\noindent\textbf{Class-weighted BCE.}
A stronger BCE baseline is class-weighted BCE, a standard strategy for imbalanced classification~\citep{he2009learning,cui2019classbalanced}, which reweights foreground and background tokens:
\begin{equation}
\mathcal{L}_{\mathrm{wBCE}}
=
-\frac{1}{N}
\sum_i
\left[
w_{\mathcal P}\hat M_i\log p_i
+
w_{\mathcal N}(1-\hat M_i)\log(1-p_i)
\right].
\end{equation}
Its logit gradient is
\begin{equation}
\frac{\partial \mathcal{L}_{\mathrm{wBCE}}}{\partial z_i}
=
\begin{cases}
-\dfrac{w_{\mathcal P}}{N}(1-p_i),
& i\in\mathcal{P},\\[6pt]
+\dfrac{w_{\mathcal N}}{N}p_i,
& i\in\mathcal{N}.
\end{cases}
\end{equation}
We use inverse-frequency class balancing,
$w_{\mathcal P}=N/(2|\mathcal P|)$ and
$w_{\mathcal N}=N/(2|\mathcal N|)$, yielding
\begin{equation}
G_{\mathcal P}
=
\frac{1}{2|\mathcal P|}
\sum_{i\in\mathcal P}(1-p_i),
\qquad
G_{\mathcal N}
=
\frac{1}{2|\mathcal N|}
\sum_{i\in\mathcal N}p_i.
\end{equation}
Thus, class-weighted BCE is a stronger token-wise baseline than vanilla BCE because it statically normalizes the foreground/background token counts.
However, its aggregate update still depends on the current prediction probabilities and does not explicitly enforce a margin between foreground and background logits.
In contrast, the proposed hinge term directly regularizes class-level logit means and provides balanced aggregate gradient mass when the margin is active.

\noindent\textbf{Focal loss.}
The standard binary focal loss~\citep{lin2017focal} is:
\begin{equation}
\mathcal{L}_{\mathrm{focal}}
=
-\frac{1}{N}
\sum_i
\left[
\alpha\,\hat M_i(1-p_i)^\gamma\log p_i
+
(1-\alpha)(1-\hat M_i)p_i^\gamma\log(1-p_i)
\right],
\end{equation}
where $\alpha\in(0,1)$ is a static foreground weighting coefficient and $\gamma$ controls the strength of hard-example modulation.
Its logit gradient is
\begin{equation}
\frac{\partial \mathcal{L}_{\mathrm{focal}}}{\partial z_i}
=
\begin{cases}
-\dfrac{\alpha}{N}(1-p_i)^\gamma
\left[(1-p_i)-\gamma p_i\log p_i\right],
& i\in\mathcal{P},\\[8pt]
+\dfrac{1-\alpha}{N}p_i^\gamma
\left[p_i-\gamma(1-p_i)\log(1-p_i)\right],
& i\in\mathcal{N}.
\end{cases}
\end{equation}
The $\alpha$ term provides static foreground--background reweighting, and the focal factors $(1-p_i)^\gamma$ and $p_i^\gamma$ suppress easy tokens.
However, focal loss is still applied independently to each token and does not impose sample-wise class-level normalization.
Its aggregate class-level gradient masses remain
\begin{equation}
G_{\mathcal P}
=
\frac{\alpha}{N}
\sum_{i\in\mathcal P}
(1-p_i)^\gamma
\left[(1-p_i)-\gamma p_i\log p_i\right],
\end{equation}
\begin{equation}
G_{\mathcal N}
=
\frac{1-\alpha}{N}
\sum_{i\in\mathcal N}
p_i^\gamma
\left[p_i-\gamma(1-p_i)\log(1-p_i)\right].
\end{equation}
Thus, the ratio $G_{\mathcal P}/G_{\mathcal N}$ still depends on the class sizes, current prediction errors, and the fixed choice of $\alpha$.
A constant $\alpha$ can reduce imbalance on average, but it cannot guarantee balanced foreground/background update pressure across images or training stages.

\noindent\textbf{Soft Dice and Tversky.}
The soft Dice loss~\citep{milletari2016vnet} is defined as
\begin{equation}
\mathcal{L}_{\mathrm{Dice}}
=
1-\frac{2A}{B},
\qquad
A=\sum_i p_i\hat M_i,
\qquad
B=\sum_i p_i+\sum_i\hat M_i .
\end{equation}
Here, $p_i=\sigma(z_i)$ is the predicted routing probability,
$\hat M_i\in\{0,1\}$ is the rasterized foreground label,
$A$ is the soft foreground overlap, and $B$ is the Dice normalization term.
Its gradient with respect to the routing logit $z_i$ is
\begin{equation}
\label{eq:dice_grad}
\frac{\partial \mathcal{L}_{\mathrm{Dice}}}{\partial z_i}
=
-\frac{2}{B^2}
\left(\hat M_i B-A\right)p_i(1-p_i)
=
\begin{cases}
-\dfrac{2(B-A)}{B^2}p_i(1-p_i),
& i\in\mathcal{P},\\[8pt]
+\dfrac{2A}{B^2}p_i(1-p_i),
& i\in\mathcal{N},
\end{cases}
\end{equation}
where $\mathcal{P}=\{i\mid \hat M_i=1\}$ and
$\mathcal{N}=\{i\mid \hat M_i=0\}$ denote foreground and background token sets, respectively.
This reveals two mechanisms that can weaken routing supervision.
First, the gradient magnitude is coupled to the global overlap statistics $A$ and $B$, which can be small or unstable when the foreground is sparse.
Second, every token is multiplied by the sigmoid factor $p_i(1-p_i)$, so saturated logits receive little correction.
For example, with $|\mathcal{P}|=12$, $|\mathcal{N}|=244$, and uniform $p_i=0.1$, we have
$A=1.2$ and $B=37.6$.
Eq.~\eqref{eq:dice_grad} then gives
\begin{equation}
G_{\mathcal{P}}
=
\sum_{i\in\mathcal{P}}
\left|
\frac{\partial \mathcal{L}_{\mathrm{Dice}}}{\partial z_i}
\right|
\approx 0.056,
\qquad
G_{\mathcal{N}}
=
\sum_{i\in\mathcal{N}}
\left|
\frac{\partial \mathcal{L}_{\mathrm{Dice}}}{\partial z_i}
\right|
\approx 0.037,
\end{equation}
which is more than an order of magnitude smaller than the unit class-level gradient mass provided by the hinge loss below.
The Tversky loss~\citep{salehi2017tversky} generalizes Dice by reweighting false positives and false negatives:
\begin{equation}
\mathcal{L}_{\mathrm{Tversky}}
=
1-
\frac{A}
{A+\alpha_T\sum_i p_i(1-\hat M_i)
+\beta_T\sum_i(1-p_i)\hat M_i},
\end{equation}
where $\alpha_T$ and $\beta_T$ control the penalties on false positives and false negatives, respectively.
Although this reweighting changes the relative cost of the two error types, the loss still inherits the global-denominator coupling and the $p_i(1-p_i)$ saturation factor.

\noindent\textbf{Margin hinge.}
Our margin hinge regularizer operates directly on class-level logit means:
\begin{equation}
\mu_{\mathcal{P}}
=
\frac{1}{|\mathcal{P}|}
\sum_{i\in\mathcal{P}}z_i,
\qquad
\mu_{\mathcal{N}}
=
\frac{1}{|\mathcal{N}|}
\sum_{i\in\mathcal{N}}z_i,
\end{equation}
\begin{equation}
\mathcal{L}_{\mathrm{hinge}}
=
\left[
m-(\mu_{\mathcal{P}}-\mu_{\mathcal{N}})
\right]_+ .
\end{equation}
When the margin is violated, i.e., $\mu_{\mathcal{P}}-\mu_{\mathcal{N}}<m$, its gradient is
\begin{equation}
\frac{\partial \mathcal{L}_{\mathrm{hinge}}}{\partial z_i}
=
\begin{cases}
-1/|\mathcal{P}|, & i\in\mathcal{P},\\[2pt]
+1/|\mathcal{N}|, & i\in\mathcal{N},
\end{cases}
\end{equation}
and is zero otherwise.
Therefore, when active,
\begin{equation}
G_{\mathcal{P}}=G_{\mathcal{N}}=1,
\end{equation}
independent of class size, current probability values, and training stage.
During optimization, the hinge term pushes foreground logits upward and background logits downward with equal aggregate magnitude, directly enlarging the foreground--background logit gap.
For numerical stability, the implementation computes the class means with denominators
$|\mathcal P|+\epsilon$ and $|\mathcal N|+\epsilon$, where $\epsilon=10^{-6}$.
If either set is empty, its summation is zero and no gradient is assigned to that empty set.
In our grounding benchmarks, each training sample contains an annotated target box, so $\mathcal P$ is normally non-empty after rasterization; this fallback is used only as a safeguard.
\begin{table}[h]
\centering
\small
\setlength{\tabcolsep}{6pt}
\caption{\textbf{Routing supervision under different imbalance losses.}
Ratio denotes average HR-token retention.
Let $\mathcal{G}$ be the ground-truth foreground token set and $\mathcal{R}$ be the set of tokens routed to HR refinement.
FG-Recall denotes $|\mathcal{R}\cap\mathcal{G}|/|\mathcal{G}|$, measuring foreground coverage, while FG-Precision denotes $|\mathcal{R}\cap\mathcal{G}|/|\mathcal{R}|$.}
\label{tab:loss_compare}
\begin{tabular}{l c c c c}
\toprule
Loss & Ratio & Overall $\uparrow$ & FG-Recall $\uparrow$ & FG-Precision $\uparrow$ \\
\midrule
Dice~\citep{milletari2016vnet}
& 33.25\% & 46.85 & 0.42 & 0.240 \\
Focal~\citep{lin2017focal} ($\alpha=0.25,\gamma=2$)
& 32.86\% & 47.82 & 0.45 & 0.260 \\
Tversky~\citep{salehi2017tversky}
& 33.38\% & 48.15 & 0.49 & 0.279 \\
Margin hinge (ours)
& 33.05\% & \textbf{49.39} & \textbf{0.54} & \textbf{0.311} \\
\bottomrule
\end{tabular}
\end{table}

\noindent\textbf{Empirical comparison.}
We retrain SmartRes-Lite with each imbalance objective while keeping all other components fixed.
Tab.~\ref{tab:loss_compare} reports the average HR-token retention ratio, overall grounding score, foreground recall, and foreground precision.
All methods use a comparable token budget of roughly $33\%$ HR-token retention.
The margin hinge achieves the best routing quality and grounding score: compared with the strongest non-hinge baseline, Tversky, it improves Overall by $+1.24$, FG-Recall by $+0.05$, and FG-Precision by $+0.032$.
\section{Attn2HR}
\label{sec:attn2hr_crop}
We introduce a simple attention-guided high-resolution selection baseline, denoted as \textit{Attn2HR}, to test whether a training-free attention heuristic is sufficient for selecting regions that require high-resolution processing.
Given the low-resolution input, \textit{Attn2HR} runs the vision encoder and extracts visual self-attention from a late visual layer. In our experiments, we use layer $\ell=30$.
Let $\mathbf A^{(\ell,h)} \in \mathbb R^{N_{\mathrm{LR}}\times N_{\mathrm{LR}}}$ denote the visual self-attention matrix of head $h$ at layer $\ell$, where $N_{\mathrm{LR}}$ is the number of valid LR visual tokens.
We compute the attention saliency score of LR token $i$ by aggregating the attention it receives from all visual tokens and heads:
\begin{equation}
s_i^{\mathrm{attn}}
=
\frac{1}{H N_{\mathrm{LR}}}
\sum_{h=1}^{H}
\sum_{j=1}^{N_{\mathrm{LR}}}
\mathbf A^{(\ell,h)}_{j,i},
\label{eq:attn2hr_score}
\end{equation}
where $H$ is the number of attention heads and $\mathbf A^{(\ell,h)}_{j,i}$ denotes the attention from query token $j$ to key token $i$.
We then select the indices of the top-$K$ LR parent tokens according to $s_i^{\mathrm{attn}}$:
\begin{equation}
\mathcal S_K
=
\operatorname{TopKIdx}
\left(
\{s_i^{\mathrm{attn}}\}_{i=1}^{N_{\mathrm{LR}}}, K
\right).
\label{eq:attn2hr_topk}
\end{equation}
Each selected LR parent token is mapped to its corresponding HR child patches using the parent-child correspondence defined in Eq.~\ref{eq:hr2lr_snap}.
The selected HR patches are then processed by the same HR re-encoding module and assembled using the same order-preserving sequence construction as Eq.~\ref{eq:Ems}.
By keeping the HR re-encoding module and sequence assembly identical to SmartRes, \textit{Attn2HR} isolates the effect of the selection policy: it replaces learned task-supervised routing with a fixed attention-based heuristic.
In Tab.~\ref{tab:ablation_strategy}, we set $K=25$, corresponding to approximately 25.5\% of LR parent tokens and yielding a final token ratio of 32.96\%.
\begin{figure}[t]
    \centering
    \includegraphics[width=\textwidth]{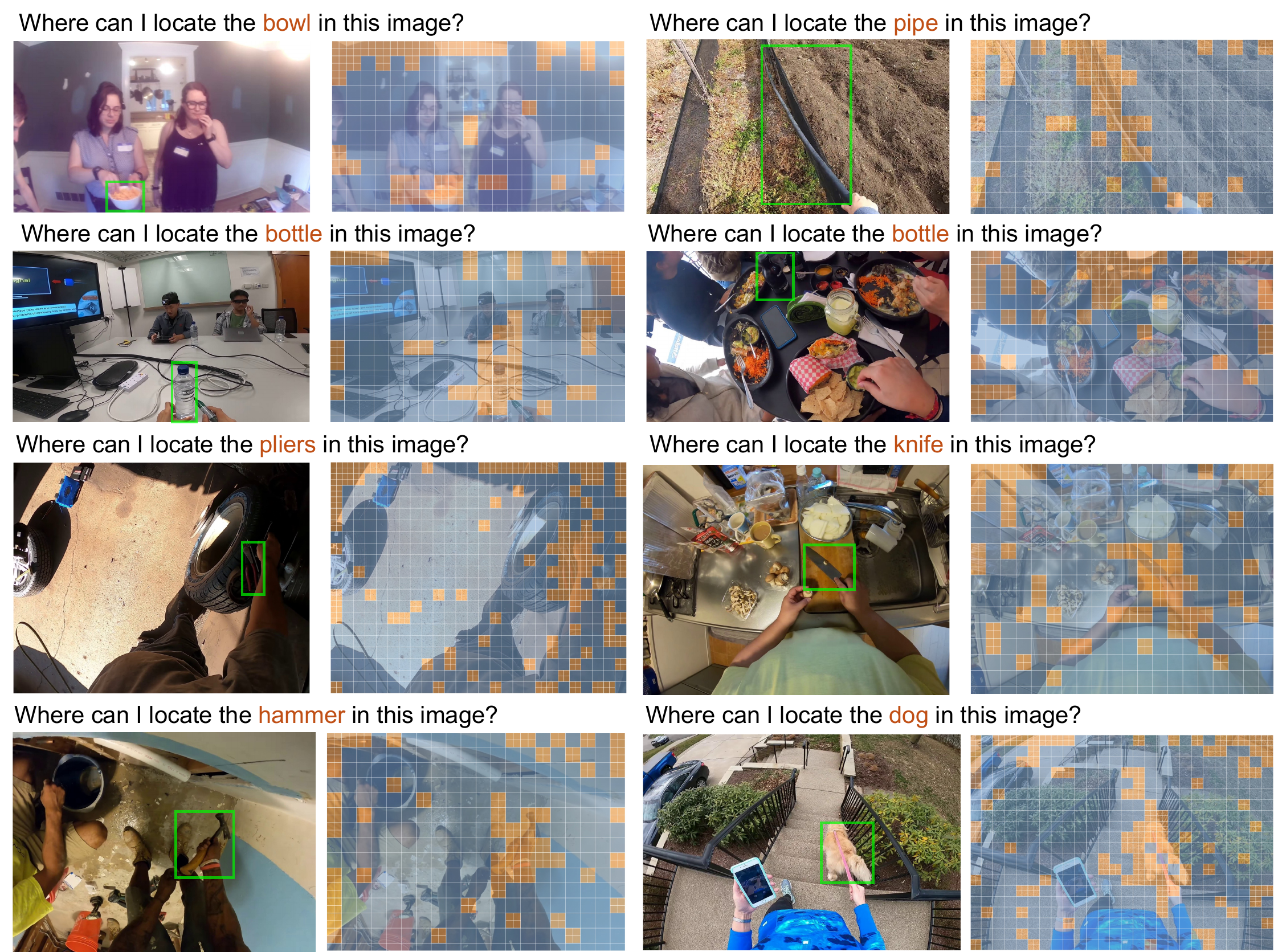}
    \caption{
    Visualizations of SmartRes' routing mask. SmartRes dynamically routes high-resolution computation (\textcolor[RGB]{244,177,131}{\rule[-0.1ex]{1.5ex}{1.5ex}}) to object-centric areas against a low-resolution background (\textcolor[RGB]{189,215,238}{\rule[-0.1ex]{1.5ex}{1.5ex}}).
   }
    \label{fig:more_visualizations}
\end{figure}
\section{Qualitative Analysis}
Additional examples of SmartRes routing are provided in Fig.~\ref{fig:more_visualizations}.
As shown, SmartRes consistently allocates HR computation to object-centric regions while keeping the surrounding scene as LR context.
This is especially beneficial for small objects, such as the \emph{pliers}, \emph{knife}, and \emph{bottle}, where fine-grained visual evidence is easily lost under uniform down-sampling or latent token pruning.
The routed regions also often cover nearby hands, tools, or manipulated objects, which provide useful spatial and interaction cues for egocentric grounding.
These examples further demonstrate that SmartRes preserves global context while selectively enhancing fine-grained details needed for accurate localization.
\section{Limitations}
SmartRes establishes patch-level resolution routing for efficient high-resolution image grounding.
A natural next direction is to extend this design from static images to video-centric egocentric perception, where routing decisions can be made temporally consistent across frames.
This would require motion-aware patch selection, temporal token reuse, and streaming memory management so that high-resolution computation is allocated not only to spatially salient regions but also to temporally persistent objects and interactions.
Another promising direction is to evaluate SmartRes in closed-loop embodied settings.
Beyond grounding metrics such as P$@0.5$ and mIoU, future studies can assess whether efficient high-resolution routing improves downstream task success, grasp accuracy, manipulation robustness, response latency, and safety-critical failure rates.
These extensions would help connect efficient visual token allocation with practical deployment in assistive technologies, robotics, and resource-constrained multimodal systems.

\end{document}